%% file: main.tex
\pdfoutput=1
\def\arxivversion{}
\documentclass[10pt,twocolumn,letterpaper]{article}

\ifdefined\arxivversion
  \usepackage[pagenumbers]{cvpr}
  \makeatletter
  \let\cvpr@maketitle\@maketitle
  \def\@maketitle{%
    \setbox\z@\vbox{\def\@author{Anonymous \confName~submission}\cvpr@maketitle}%
    \vbox to\ht\z@{%
      \null\vskip 10pt
      \begin{center}
        {\Large\bf\@title\par}
        \vspace*{18pt}
        {\large\lineskip.5em\begin{tabular}[t]{c}\@author\end{tabular}\par}
      \end{center}
      \vss}}
  \makeatother
\else
  \usepackage[review]{cvpr}
\fi

\input{preamble}

\definecolor{cvprblue}{rgb}{0.21,0.49,0.74}
\definecolor{cvprred}{HTML}{CB181D}

\def\confName{CVPR}
\def\confYear{2027}
\AtEndPreamble{%
  \crefname{appendix}{Supp.}{Supp.}\Crefname{appendix}{Supp.}{Supp.}%
  \crefname{subappendix}{Supp.}{Supp.}\Crefname{subappendix}{Supp.}{Supp.}%
  \crefname{subsubappendix}{Supp.}{Supp.}\Crefname{subsubappendix}{Supp.}{Supp.}}
\usepackage{url}
\usepackage{booktabs}
\usepackage{amsfonts}
\usepackage{nicefrac}
\usepackage{microtype}
\usepackage{xcolor}
\usepackage{lipsum}

\definecolor{dblue}{RGB}{52, 104, 192}
\definecolor{dgreen}{RGB}{65, 171, 93}
\definecolor{dred}{RGB}{210, 69, 69}
\definecolor{dpurple}{RGB}{148, 103, 189}
\definecolor{dorange}{RGB}{255, 127, 14}
\definecolor{dcyan}{RGB}{35, 70, 100}

\definecolor{color_cem}{RGB}{102, 194, 165}
\definecolor{color_ecbm}{RGB}{252, 141, 98}
\definecolor{color_ar}{RGB}{141, 160, 203}
\definecolor{color_scbm}{RGB}{231, 138, 195}

\usepackage{bbm}
\usepackage{dsfont}
\usepackage{graphicx}
\usepackage{amsmath}
\usepackage{amssymb}
\usepackage{amsthm}
\usepackage{multirow}
\usepackage{makecell}
\usepackage{caption}
\usepackage{wrapfig}
\usepackage{enumitem}
\usepackage{float}

\usepackage[pagebackref,breaklinks,colorlinks,allcolors=cvprblue,linkcolor=cvprred]{hyperref}
\usepackage{subcaption}
\usepackage{multicol}
\usepackage{colortbl}
\colorlet{tabgray}{gray!12}
\colorlet{tabhead}{gray!25}
\newcommand{\gc}{\cellcolor{tabgray}}
\newcommand{\tpm}[1]{\,{\scriptsize$\pm$#1}}
\newlength{\alnumL}\newlength{\alnumR}
\newcolumntype{A}[2]{>{\settowidth{\alnumL}{$#1$}\settowidth{\alnumR}{\tpm{#2}}\centering\arraybackslash}X}
\newcolumntype{N}[2]{>{\settowidth{\alnumL}{$#1$}\settowidth{\alnumR}{\tpm{#2}}}c}
\NewDocumentCommand{\alignednum}{m o}{%
  \makebox[\alnumL][r]{#1}\makebox[\alnumR][l]{\IfValueT{#2}{\textcolor{black!55}{\tpm{#2}}}}}
\usepackage{xspace}
\usepackage{siunitx}
\usepackage{array}
\usepackage{tabularx}
\newcolumntype{Y}{>{\centering\arraybackslash}X}
\usepackage{pifont}
\usepackage{placeins}


\input{symbols}

\input{math_commands}

\title{Beyond Rigid Hands: Deforming Skin into Contact-Consistent Hand Meshes}

\ifdefined\arxivversion
\author{Seonghwan Park \qquad Min-Gyu Park\\
Korea Electronics Technology Institute (KETI)\\
{\tt\small \{shpark97, mpark\}@keti.re.kr}
}
\else
\author{Anonymous Authors}
\fi

\begin{document}
\maketitle
\input{sec/0_Abstract}    
\input{sec/1_Introduction}
\input{sec/2_Related_work}
\input{sec/3_Method}
\input{sec/4_Evaluation}
\input{sec/5_Conclusion}

{
    \small
    \bibliographystyle{ieeenat_fullname}
    \bibliography{main}
}

\clearpage
\appendix
\crefalias{section}{appendix}\crefalias{subsection}{subappendix}\crefalias{subsubsection}{subsubappendix}
\input{sec/6_Supplementary}

\end{document}

%% file: preamble.tex
\usepackage{xspace}

\usepackage{newtxtext}

%% file: symbols.tex
\newcommand{\cmark}{\textcolor{dblue}{\ding{51}}}%

\definecolor{tabgray}{HTML}{f2f2f2}
\definecolor{tablavender}{HTML}{E6F7E6}
\definecolor{tabnavy}{HTML}{00008B}
\definecolor{tabwhite}{HTML}{FFFFFF}
\definecolor{linf}{HTML}{8172b2}
\definecolor{l2}{HTML}{ccb974}
\definecolor{l1}{HTML}{64b5cd}

\definecolor{lavender}{HTML}{9ad756}
\definecolor{DBlue}{RGB}{31, 119, 180}
\definecolor{DGreen}{RGB}{44,160,44}
\definecolor{DRed}{RGB}{214,39,40}
\definecolor{dorange}{RGB}{255, 127, 14}
\definecolor{dorange2}{RGB}{255, 165, 0}

%% file: math_commands.tex
\usepackage{bm, dsfont}

\let\hat\widehat

\newcommand{\RR}{\mathbb{R}}

\makeatletter
\newcommand{\ve}{\@ifnextchar\bgroup{\velong}{{\bm{e}}}}
\newcommand{\velong}[1]{{\bm{#1}}}
\makeatother

%% file: sec/0_Abstract.tex
\begin{abstract}
Hands grasp objects with soft skin.
A fingertip flattens against a mug, and a palm curves around a bottle.
Yet hand-object datasets and estimators represent the hand as a rigid articulated mesh, which hovers above or penetrates the object where skin would press against it.
Existing refiners keep the mesh rigid and adjust only the pose, trading pose accuracy for physical plausibility.
We ask whether a rigid hand can instead touch the object as skin does while staying close to its original pose.
We lift annotated or estimated rigid hands to contact-consistent meshes through an optimization that (i) slightly adjusts the joints and wrist to bring the fingers onto the object, (ii) deforms the skin so that penetrating regions settle onto the surface, and (iii) keeps this deformation consistent over time.
Since deformation reaches only skin already near the object, we further train a lightweight head on the lifted annotations to predict where an estimated hand should touch and steer the deformation there.
Unlike existing refiners, the lift more than doubles the skin lying on the object surface for three published estimators while barely changing the pose.
Guidance from the learned head further improves grasp stability by closing contact gaps that the estimate leaves open.
Existing hand-object datasets and estimators can thus adopt contact-consistent hands without changing the hand model or retraining the estimators.
\end{abstract}

%% file: sec/1_Introduction.tex
\addtocontents{toc}{\protect\setcounter{tocdepth}{0}}
\section{Introduction}
\label{sec:introduction}

Reconstructing how a hand grasps an object in 3D is central to understanding manipulation~\cite{hasson2019obman,fan2023arctic}.
Large hand-object datasets~\cite{chao2021dexycb,yang2022oakink,hampali2020honnotate,liu2022hoi4d,fan2023arctic} annotate hundreds of thousands of grasp frames, and estimators trained on them reconstruct hands and objects from images~\cite{qi2024hoisdf,wang2025unihope,zhou2026vpho}.
Both represent the hand with MANO~\cite{romero2017mano}, a compact parametric model whose skinned mesh is fully determined by joint angles and shape parameters.
The hand therefore articulates at its joints but its surface stays rigid, which suffices for describing pose.
Yet a grasp is decided at \emph{contact}, which determines whether the object is held stably~\cite{murray1994mathematical} and is what contact estimators learn to predict~\cite{tse2022s2contact,jung2025haco}.
Even at an accurate pose, a rigid hand misrepresents this contact because its surface rarely meets the object the way real skin does.

\begin{figure}[t]
  \centering
  \includegraphics[width=\linewidth]{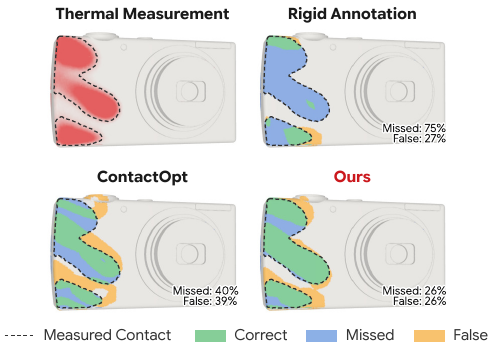}
  \caption{Rigid hands misrepresent contact.
  The rigid annotation misses most of the thermally measured contact~\cite{brahmbhatt2020contactpose}, the pose refiner ContactOpt~\cite{grady2021contactopt} adds false contact, and our lift covers more of the measurement while staying near the annotated pose.
  \emph{Missed} is measured contact left uncovered, and \emph{False} is contact outside it.}
  \label{fig:teaser}
\end{figure}

\begin{figure*}[t]
  \centering
  \begin{subfigure}[b]{0.645\linewidth}
    \centering
    \includegraphics[width=\linewidth]{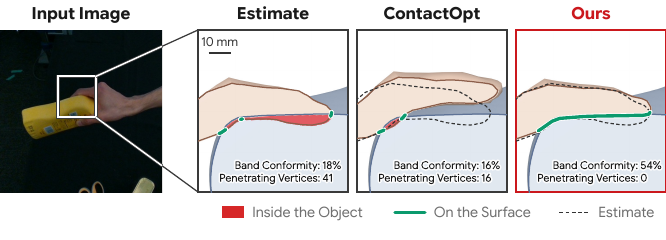}
    \caption{Contact at the estimated pose}
    \label{fig:posthoc_frame}
  \end{subfigure}
  \hfill
  \begin{subfigure}[b]{0.334\linewidth}
    \centering
    \includegraphics[width=\linewidth]{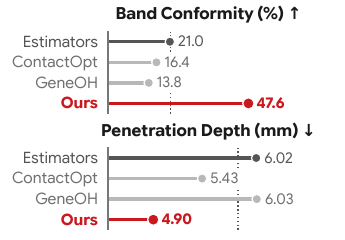}
    \caption{Comparison with pose refiners}
    \label{fig:posthoc_chart}
  \end{subfigure}
  \caption{Refining estimated hands by deforming the skin rather than changing the pose.
  (a) A finger predicted by an estimator~\cite{zhou2026vpho} penetrates the object (\textcolor[HTML]{D62728}{red}).
  A pose refiner~\cite{grady2021contactopt} moves the finger away from the estimate (dashed) yet leaves its base inside, whereas our lift stays at the estimate and brings the skin onto the surface (\textcolor[HTML]{0B9A6D}{green}).
  Each panel reports band conformity (the share of near-contact skin lying on the surface) and the number of penetrating vertices.
  (b) Our lift more than doubles the average band conformity of three estimators~\cite{qi2024hoisdf,wang2025unihope,zhou2026vpho} and reduces penetration depth the most, while pose refiners~\cite{grady2021contactopt,liu2024geneoh} lower conformity.
  Dotted lines mark the rigid ground-truth annotations.}
  \label{fig:posthoc_overview}
\end{figure*}

The limitation lies in the representation itself.
Skin under grip deforms into patches that conform to the object~\cite{serina1997force}, whereas a rigid surface can only hover above or penetrate it.
As a result, even ground-truth annotations miss most of the contact recorded by thermal cameras~\cite{brahmbhatt2020contactpose} (\Cref{fig:teaser}).
Existing refiners~\cite{grady2021contactopt,yang2021cpf,zhou2022toch,liu2024geneoh} keep the surface rigid and adjust the pose instead, which moves the hand away from its annotated or estimated placement without making it conform.
For instance, freeing a fingertip can leave the finger's base inside the object (\Cref{fig:posthoc_overview}).
What is missing is not a better pose but a surface that can deform.
This motivates our central question: \emph{Can the skin of an annotated or estimated rigid hand conform to the object while its pose barely changes?}

Answering this question raises two challenges.
First, deformed hands have no ground truth to validate a lift against.
Contact alone is insufficient, since simply snapping the skin onto the object also creates contact yet stretches the skin, departs from the observed depth, and flickers between frames.
A credible lift must avoid these failures and be verified against data the optimization never sees, such as depth images~\cite{chao2021dexycb} and thermal contact maps~\cite{brahmbhatt2020contactpose}.
Second, geometry alone cannot tell whether a finger should touch when an imprecise estimate leaves it short of the object.
This calls for a learned contact cue, which rigid annotations cannot reliably supervise.

To this end, we lift a rigid hand in three stages that create contact while avoiding the failures above:
(i)~\textbf{kinematic seating} brings the fingers onto the object by slightly bending or straightening each joint about its original axis and shifting the wrist;
(ii)~\textbf{skin deformation} then moves individual vertices so that hovering and penetrating skin settle onto the surface while a regularizer on per-face deformation gradients~\cite{NJF} lets palm-side skin near the object yield but keeps the back of the hand stiff to prevent stretching;
and (iii)~a \textbf{temporal prior} couples the frames of a sequence by penalizing changes of each displacement in the rest frame of its bones so that an indentation persists as the fingers move instead of flickering.
Since the lifted mesh keeps the MANO topology, the same lift can replace the rigid hands in dataset annotations and refine the outputs of published estimators without retraining them.
Yet geometry reveals where the skin of an estimated hand already meets the object but not where it should, so we further train a lightweight contact head on the lifted annotations.
The head maps the features of the frozen estimator to the distance from each vertex to the object, and skin deformation then pulls the vertices predicted to be in contact onto the surface.

Our contributions are threefold:
\begin{itemize}[leftmargin=1em, topsep=-\parskip, itemsep=0.1em]
    \item We propose an optimization that lifts annotated or estimated rigid hands to contact-consistent meshes through seating, skin deformation, and a temporal prior, increasing skin contact with minimal pose change.
    \item We train a feed-forward contact head on lifted annotations to predict where estimated hands should touch, and it yields more stable grasps than the same head trained on rigid annotations.
    \item We validate the lift on data the optimization never sees, showing that lifted annotations stay consistent with observed depth while holding objects more stably, matching thermal contact better, and training more accurate contact estimators than rigid annotations.
\end{itemize}

%% file: sec/2_Related_work.tex
\section{Related Work}
\label{sec:related_work}

\noindent\textbf{Rigid hands in data and estimation.}\enspace
Hand-object datasets annotate grasps by fitting rigid MANO hands~\cite{romero2017mano} to images, depth, or markers~\cite{chao2021dexycb,yang2022oakink,hampali2020honnotate,liu2022hoi4d,fan2023arctic}, and models trained on them inherit this representation in hand and hand-object reconstruction~\cite{pavlakos2024hamer,potamias2024wilor,hasson2019obman,lin2023hflnet,wang2025unihope,liu2025easyhoi,wang2025magichoi} and in contact estimation~\cite{jung2025haco,fan2023arctic}.
Their contact is therefore read off a surface that cannot deform and carries the gaps and penetrations of the rigid fit.
Thermal imaging~\cite{brahmbhatt2020contactpose} and pressure sensing~\cite{grady2022pressurevision,zhao2025egopressure} measure contact independently of any mesh and can validate such annotations.
We lift these annotations, learn contact from them, and verify both against thermal imprints.

\noindent\textbf{Contact refinement through pose.}\enspace
Refiners make rigid hands more plausible by optimizing the pose toward predicted contact~\cite{grady2021contactopt,yang2021cpf} or by denoising hand trajectories with learned priors~\cite{zhou2022toch,liu2024geneoh}, while physics-based and generative methods add force, stability, or contact-consistency objectives~\cite{hu2022physical,luo2024physics,liu2021synthesizing,wang2023dexgraspnet,jiang2021graspTTA}.
All of them keep the mesh rigid and treat skin compliance at most as a penetration allowance~\cite{grady2021contactopt}, so they create contact by moving the hand at the cost of the observed pose.
By contrast, our lift keeps the pose nearly fixed and lets the skin deform.

\begin{figure*}[t]
  \centering
  \begin{subfigure}[b]{0.3636\linewidth}
    \centering
    \includegraphics[width=\linewidth]{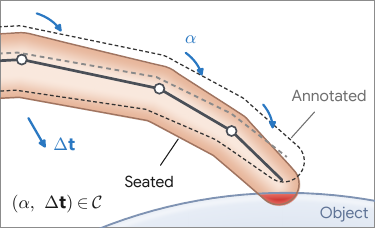}
    \caption{Kinematic seating}
    \label{fig:method_seating}
  \end{subfigure}
  \hfill
  \begin{subfigure}[b]{0.2909\linewidth}
    \centering
    \includegraphics[width=\linewidth]{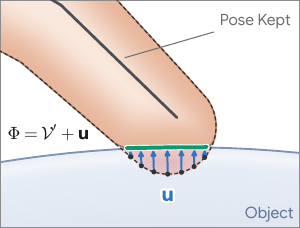}
    \caption{Skin deformation}
    \label{fig:method_deform}
  \end{subfigure}
  \hfill
  \begin{subfigure}[b]{0.3055\linewidth}
    \centering
    \includegraphics[width=\linewidth]{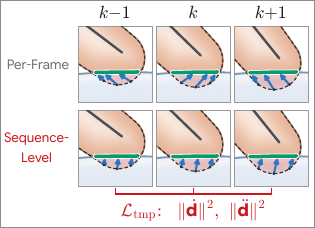}
    \caption{Sequence-level optimization}
    \label{fig:method_temporal}
  \end{subfigure}
  \caption{Lifting a rigid hand in three stages.
  (a)~Seating bends the annotated finger (dashed) toward the object by scaling its joint angles by $\alpha$ and shifting the wrist by $\Delta\mathbf{t}$ within the box $\mathcal{C}$, so that it presses slightly into the object (\textcolor[HTML]{D62728}{red}).
  (b)~The displacement $\mathbf{u}$ pushes the penetrating skin back onto the surface as a flat contact patch (\textcolor[HTML]{0B9A6D}{green}) without changing this pose, while the Jacobian regularizer keeps the skin from stretching.
  (c)~Per-frame lifting changes the patch between frames although the finger barely moves (top), whereas the temporal prior $\mathcal{L}_{\mathrm{tmp}}$ on the canonical deformation $\mathbf{d}$ keeps it consistent (bottom).}
  \label{fig:method}
\end{figure*}

\noindent\textbf{Deformable hands.}\enspace
Anatomical hand models with bones, muscles, or soft tissue~\cite{li2022nimble,sun2024physhand} cannot be directly applied to existing MANO annotations.
Graphics methods deform skinned characters under contact through elasticity~\cite{mcadams2011efficient,chao2010simple} or implicit skinning~\cite{vaillant2013implicit} and track or learn the soft tissue of captured hands and bodies~\cite{smith2020constraining,santesteban2020softsmpl}, but they animate or capture motion rather than preserve an annotated pose.
Deformable reconstruction from video lets objects~\cite{zhong2024springgaus,jiang2025phystwin,zhang2024gsdynamics,lee2026physhandi} or faces~\cite{shimada2023decaf,he2025headcontact} deform while the hand stays rigid.
We deform the MANO skin itself and regularize per-face deformation gradients~\cite{NJF} toward the identity, so the lifted hand remains a drop-in replacement for the rigid one.

%% file: sec/3_Method.tex
\section{Lifting and Learning Contact}
\label{sec:dataset}

Given a rigid MANO hand and a posed object, we first define the lift and the contact consistency it targets (\Cref{sec:dataset_problem}).
We then seat the hand to bring the fingers onto the object (\Cref{sec:dataset_kin}), deform the skin to conform to it without stretching (\Cref{sec:dataset_method}), and couple frames to keep the deformation consistent over time (\Cref{sec:dataset_temporal}).
Finally, we learn a head that predicts the lifted contact from frozen estimator features and guides the lift of estimated hands (\Cref{sec:method_learn}).
\Cref{fig:method} illustrates the three lifting stages.

\subsection{Problem Definition}
\label{sec:dataset_problem}

We build on MANO~\cite{romero2017mano}, which maps pose $\theta \in \RR^{48}$ and shape $\beta \in \RR^{10}$ to a hand mesh $\mathcal{M}(\theta, \beta) \in \RR^{N \times 3}$ of $N = 778$ vertices through linear blend skinning~\cite{lewis2000pose}.
A rigid hand at wrist position $\mathbf{t} \in \RR^{3}$ is then \mbox{$\mathcal{V} = \mathcal{M}(\theta, \beta) + \mathbf{t}$}, and the object is given by its signed distance function $\Omega$~\cite{park2019deepsdf}.
The lift keeps the shape and corrects the hand in two steps: a slight pose adjustment followed by a skin displacement,
\begin{equation}
  \Phi = \underbrace{\mathcal{M}(\theta', \beta) + \mathbf{t}'}_{\mathcal{V}'} +\; \mathbf{u},
  \label{eq:lifted}
\end{equation}
where the seated pose $(\theta', \mathbf{t}')$ stays near the input and brings the fingers onto the object, $\mathcal{V}'$ is the seated hand, and the per-vertex displacement $\mathbf{u} \in \RR^{N \times 3}$ deforms its skin into the lifted surface $\Phi$ without changing the MANO topology.

We seek corrections that let $\Phi$ touch the object as skin does while keeping the hand near its input pose.
We formalize this goal as \emph{contact consistency}: the skin must not penetrate the object (C1), the skin near the object must lie on its surface (C2), and the pose correction must stay small (C3),
\begin{align}
  \text{(C1)}\quad & \phi(\Phi_i) \leq \epsilon, \quad \forall i \notin \mathcal{X}, \label{eq:c1} \\
  \text{(C2)}\quad & |\Omega(\Phi_i)| \leq \tau, \quad \forall i \in \mathcal{B}, \label{eq:c2} \\
  \text{(C3)}\quad & (\alpha, \Delta\mathbf{t}) \in \mathcal{C}, \label{eq:c3}
\end{align}
where $\Phi_i$ is the $i$-th vertex of $\Phi$, $\phi = [-\Omega]_{+}$ is the penetration depth, and $\epsilon$ and $\tau$ are small tolerances.
C1 exempts the vertices $\mathcal{X}$ that the input already places deep inside the object, since fixing such inherited errors would distort the hand.
C2 applies to the contact band $\mathcal{B}$, the palmar vertices near the object.
C3 confines the pose correction to a box $\mathcal{C}$, where $\alpha$ scales each joint rotation to slightly bend or straighten the finger and $\Delta\mathbf{t}$ shifts the wrist.

\subsection{Kinematic Seating}
\label{sec:dataset_kin}

Rigid inputs often leave the fingers slightly above the object, beyond the reach of skin deformation.
Seating first closes such gaps with a small pose correction (\Cref{fig:method_seating}).
Unlike pose refiners that optimize the full pose~\cite{grady2021contactopt,yang2021cpf}, it scales only the angle of each finger joint $j$ and shifts the wrist,
\begin{equation}
  \label{eq:kin_seat}
  \theta'_j = (1 + \alpha_j)\, \theta_j, \qquad \mathbf{t}' = \mathbf{t} + \Delta\mathbf{t},
\end{equation}
where $\theta_j$ is the axis-angle rotation of joint $j$ and $\Delta\mathbf{t} \in \RR^{3}$ translates the whole hand toward the object.
Scaling preserves the input rotation axis, so each joint has a single degree of freedom that bends or straightens it without twist.
We optimize $(\alpha, \Delta\mathbf{t})$ with Adam~\cite{kingma2014adam}, projecting onto the box $\mathcal{C}$ of C3 after each step,
\begin{equation}
  \label{eq:kin_obj}
  \min_{(\alpha,\, \Delta\mathbf{t}) \in \mathcal{C}}\ w_{\mathrm{att}}\, \mathcal{L}'_{\mathrm{att}} + w_{\mathrm{rep}}\, \mathcal{L}'_{\mathrm{rep}} + w_{\mathrm{prior}}\, \mathcal{L}_{\mathrm{prior}}.
\end{equation}
The attraction $\mathcal{L}'_{\mathrm{att}}$ pulls the palmar skin near the object slightly inside its surface and the repulsion $\mathcal{L}'_{\mathrm{rep}}$ penalizes deeper penetration as in hand-object contact losses~\cite{hasson2019obman,grady2021contactopt}, while the prior $\mathcal{L}_{\mathrm{prior}}$ keeps the correction small.
Distant fingers are not attracted and keep their input angles.
Seating alone barely improves conformity (C2), since the rigid surface still hovers above or penetrates the object around each contact.
We therefore seat the fingers slightly into the object, letting skin deformation flatten the penetrating skin onto its surface.
The full loss definitions are given in \Cref{app:setup}.

\subsection{Skin Deformation}
\label{sec:dataset_method}

Seating places the fingers but leaves the rigid surface unchanged.
Skin deformation makes this surface conform to the object by displacing the seated vertices by $\mathbf{u}$ (\Cref{fig:method_deform}).
Simply projecting them onto the object would stretch the skin, so we draw on Aigerman~\etal~\cite{NJF} and regularize the deformation through per-face Jacobians.
The Jacobian $J_f \in \RR^{3 \times 3}$ of face $f$ is its deformation gradient~\cite{sumner2004deformation} and linearly maps the face from the seated to the lifted hand,
\begin{equation}
  \label{eq:jac}
  J_f = I + \nabla_{\!f}\, \mathbf{u},
\end{equation}
where the gradient $\nabla_{\!f}\, \mathbf{u}$ of the displacement over the face captures how $\mathbf{u}$ differs between its three vertices.
$J_f$ is thus the identity $I$ when the three vertices move together and departs from it as the face stretches, shears, or rotates, so we penalize this departure over the faces $\mathcal{F}$,
\begin{equation}
  \label{eq:reg}
  \mathcal{L}_{\mathrm{reg}} = \frac{1}{|\mathcal{F}|} \sum_{f \in \mathcal{F}} m_f \big\| J_f - I \big\|_F^2.
\end{equation}
Since the penalty ignores translation, a skin patch can move onto the surface as a whole while resisting stretch.
The stiffness $m_f$ is relaxed on palmar faces near the object to let the skin pressing on it yield, and raised on dorsal faces to keep the back of the hand stiff.
With contact terms for C1 and C2, the displacement minimizes
\begin{equation}
  \label{eq:dataset_loss}
  \mathcal{L}(\mathbf{u}) =
  \underbrace{\lambda_{\mathrm{att}} \mathcal{L}_{\mathrm{att}} + \lambda_{\mathrm{pen}} \mathcal{L}_{\mathrm{pen}}}_{\text{contact}}
  + \underbrace{\lambda_{\mathrm{reg}} \mathcal{L}_{\mathrm{reg}} + \lambda_{\mathrm{ver}} \mathcal{L}_{\mathrm{ver}}}_{\text{deformation}},
\end{equation}
where the attraction $\mathcal{L}_{\mathrm{att}}$ and penetration $\mathcal{L}_{\mathrm{pen}}$ mirror the seating terms but draw the skin onto the surface itself rather than slightly inside, and the cap $\mathcal{L}_{\mathrm{ver}}$ softly bounds how far each vertex moves.
Since the pose stays fixed, all contact gained here comes from the skin rather than from moving the hand.
The full loss definitions are given in \Cref{app:setup}.

\subsection{Sequence-Level Optimization}
\label{sec:dataset_temporal}

The stages above lift each frame independently and leave the skin free to slide along the object, so nearby frames settle into different local minima and the indentation flickers even when the grasp barely changes (\Cref{fig:method_temporal}).
We therefore optimize all frames of a clip jointly and couple their deformations.

Since $\mathbf{u}$ rotates with the finger that carries it, we undo this rotation before comparing frames,
\begin{equation}
  \label{eq:delta_cano}
  \mathbf{d}_{k,i} = A_{k,i}^{-1}\, \mathbf{u}_{k,i},
\end{equation}
where $\mathbf{u}_{k,i}$ is the displacement of vertex $i$ in frame $k$ of a clip of $T$ frames and $A_{k,i}$ is the blend of bone rotations that skinning applies to this vertex at the seated pose.
The canonical deformation $\mathbf{d}_{k,i}$ describes the displacement as if the hand were at rest, so an indentation that stays on the same part of the finger keeps the same $\mathbf{d}_{k,i}$ while the finger bends.

As grasp deformation varies slowly relative to the frame rate, we penalize the velocity and acceleration of $\mathbf{d}$ following temporal priors in video pose estimation~\cite{kocabas2020vibe,choi2021tcmr,rempe2021humor},
\begin{equation}
  \label{eq:findiff}
  \dot{\mathbf{d}}_k = \mathbf{d}_{k+1} - \mathbf{d}_{k}, \qquad
  \ddot{\mathbf{d}}_k = \mathbf{d}_{k+1} - 2\mathbf{d}_{k} + \mathbf{d}_{k-1},
\end{equation}
\begin{equation}
  \label{eq:temporal}
  \mathcal{L}_{\mathrm{tmp}} = \frac{\lambda_{\mathrm{vel}}}{N |\mathcal{K}_1|} \sum_{k \in \mathcal{K}_1} \big\| \dot{\mathbf{d}}_k \big\|_F^2
  + \frac{\lambda_{\mathrm{acc}}}{N |\mathcal{K}_2|} \sum_{k \in \mathcal{K}_2} \big\| \ddot{\mathbf{d}}_k \big\|_F^2 ,
\end{equation}
where $\lambda_{\mathrm{vel}}$ and $\lambda_{\mathrm{acc}}$ weight the two terms, and $\mathcal{K}_1$ and $\mathcal{K}_2$ hold consecutive pairs and triples of contact frames so that deformation stays out of frames before contact.
The clip is then optimized as
\begin{equation}
  \label{eq:seq_obj}
  \min_{\{\mathbf{u}_k\}_{k=1}^{T}}\ \sum_{k=1}^{T} \Big[ \mathcal{L}(\mathbf{u}_k) + \lambda_{\mathrm{wrist}}\, \mathcal{L}_{\mathrm{wrist}}(\mathbf{u}_k) \Big] + \mathcal{L}_{\mathrm{tmp}},
\end{equation}
where $\mathcal{L}_{\mathrm{wrist}}$ anchors the open wrist boundary of the mesh.

\begin{table*}[t]
  \centering
  \caption{Refining estimated hands on the DexYCB~\cite{chao2021dexycb} dataset.
  Each refiner receives the hand and object predicted by a published estimator~\mbox{\cite{qi2024hoisdf,wang2025unihope,zhou2026vpho}}, and our lift uses either geometry alone or the learned head.
  Hand and Temporal entries show the change from the estimate to the refined result, and C2 is measured on the band of the estimated hand.
  \textbf{Bold} and \underline{underline} mark the best and second-best results.
  For Hand and Temporal the smallest change is best, and elsewhere only refiners that improve over the estimate are marked.}
  \label{tab:posthoc}
  \vspace{-0.75em}
  \small
  \resizebox{\textwidth}{!}{%
  \begin{tabular}{l|cc|cccc|cc|c}
    \toprule
    & \multicolumn{2}{c|}{\textbf{Hand}} & \multicolumn{4}{c|}{\textbf{Contact}} & \multicolumn{2}{c|}{\textbf{Stability}} & \textbf{Temporal} \\
    \cmidrule(lr){2-3}\cmidrule(lr){4-7}\cmidrule(lr){8-9}\cmidrule(l){10-10}
      & MPJPE & MPVPE
      & P.D. & I.V. & $N_{\mathrm{in}>1}$ & C2
      & S.D.@6 & F.E.@6
      & Accel \\
    \textbf{Method}
      & {\small (mm) $\downarrow$} & {\small (mm) $\downarrow$}
      & {\small (mm) $\downarrow$} & {\small (cm$^3$) $\downarrow$} & {\small (\#) $\downarrow$} & {\small (\%) $\uparrow$}
      & {\small (cm) $\downarrow$} & {\small (\%) $\downarrow$}
      & {\small (mm/f$^2$) $\downarrow$} \\
    \midrule
    HOISDF~\cite{qi2024hoisdf} & 10.3 & 10.1 & 6.44 & 7.57 & 57.0 & 20.7 & 2.31 & 18.0 & 4.70 \\
    \quad + ContactOpt~\cite{grady2021contactopt} & 10.3\ $\to$\ 15.5 & 10.1\ $\to$\ 15.1 & 5.70 & 6.48 & \textbf{17.3} & 16.9 & \underline{2.04} & \underline{11.8} & 4.70\ $\to$\ 18.27 \\
    \quad + GeneOH Diffusion~\cite{liu2024geneoh} & 10.3\ $\to$\ 17.1 & 10.1\ $\to$\ 17.1 & 6.58 & 7.94 & 65.1 & 15.3 & 2.23 & 14.3 & 4.70\ $\to$\ 21.68 \\
    \rowcolor{tabgray} \quad + Ours & \textbf{10.3}\ $\to$\ \textbf{10.5} & \textbf{10.1}\ $\to$\ \textbf{10.3} & \textbf{5.13} & \textbf{5.98} & 18.2 & \underline{48.6} & 2.19 & 16.1 & \textbf{4.70}\ $\to$\ \phantom{0}\textbf{4.73} \\
    \rowcolor{tabgray} \quad + Ours with Head & \underline{10.3\ $\to$\ 10.7} & \underline{10.1\ $\to$\ 10.5} & \textbf{5.15} & \underline{6.36} & \underline{17.4} & \textbf{51.7} & \textbf{1.86} & \textbf{11.1} & \underline{4.70\ $\to$\ \phantom{0}4.76} \\
    \midrule
    UniHOPE~\cite{wang2025unihope} & 11.0 & 10.7 & 5.06 & 6.28 & 48.4 & 20.0 & 5.29 & 46.2 & 6.09 \\
    \quad + ContactOpt~\cite{grady2021contactopt} & 11.0\ $\to$\ 26.6 & 10.7\ $\to$\ 25.7 & 4.79 & \textbf{4.95} & \textbf{11.8} & 13.3 & \textbf{4.55} & \textbf{34.1} & 6.09\ $\to$\ 44.99 \\
    \quad + GeneOH Diffusion~\cite{liu2024geneoh} & 11.0\ $\to$\ 31.4 & 10.7\ $\to$\ 31.0 & 4.53 & 5.53 & 58.4 & 11.4 & 5.46 & 45.5 & 6.09\ $\to$\ 23.80 \\
    \rowcolor{tabgray} \quad + Ours & \textbf{11.0}\ $\to$\ \textbf{11.1} & \textbf{10.7}\ $\to$\ \textbf{10.9} & \textbf{4.39} & \underline{5.47} & 23.4 & \underline{47.5} & 5.27 & 46.5 & \textbf{6.09}\ $\to$\ \phantom{0}\textbf{6.15} \\
    \rowcolor{tabgray} \quad + Ours with Head & \underline{11.0\ $\to$\ 11.2} & \textbf{10.7}\ $\to$\ \textbf{10.9} & \textbf{4.36} & 5.56 & \underline{22.8} & \textbf{49.6} & \underline{5.18} & \underline{44.4} & \underline{6.09\ $\to$\ \phantom{0}6.25} \\
    \midrule
    VPHO~\cite{zhou2026vpho} & 8.8 & 8.6 & 6.55 & 8.23 & 48.4 & 22.3 & 1.54 & 9.3 & 5.11 \\
    \quad + ContactOpt~\cite{grady2021contactopt} & \phantom{0}8.8\ $\to$\ 12.9 & \phantom{0}8.6\ $\to$\ 12.5 & 5.79 & 7.47 & 20.0 & 19.0 & \underline{1.49} & \underline{8.6} & 5.11\ $\to$\ 17.05 \\
    \quad + GeneOH Diffusion~\cite{liu2024geneoh} & \phantom{0}8.8\ $\to$\ 14.8 & \phantom{0}8.6\ $\to$\ 15.0 & 6.98 & 8.99 & 61.2 & 14.8 & 1.50 & \textbf{7.0} & 5.11\ $\to$\ 18.26 \\
    \rowcolor{tabgray} \quad + Ours & \phantom{0}\textbf{8.8}\ $\to$\ \phantom{0}\textbf{9.1} & \phantom{0}\textbf{8.6}\ $\to$\ \phantom{0}\textbf{8.9} & \textbf{5.17} & \textbf{6.24} & \textbf{13.0} & \underline{46.7} & 1.53 & 9.3 & \textbf{5.11}\ $\to$\ \phantom{0}\textbf{5.11} \\
    \rowcolor{tabgray} \quad + Ours with Head & \phantom{0}\underline{8.8\ $\to$\ \phantom{0}9.2} & \phantom{0}\underline{8.6\ $\to$\ \phantom{0}9.0} & \textbf{5.19} & \underline{6.51} & \underline{13.5} & \textbf{49.3} & \textbf{1.39} & \textbf{7.0} & \underline{5.11\ $\to$\ \phantom{0}5.12} \\
    \bottomrule
  \end{tabular}}
\end{table*}

\subsection{Learning Contact from Lifted Annotations}
\label{sec:method_learn}

Geometry reveals where skin already meets the object but not where it should, so the gap that an estimate leaves between a finger and the object stays open.
Since rigid annotations hover above or penetrate the object, we learn this cue from the lifted annotations instead and let skin deformation also pull the vertices predicted to touch onto the surface.

We realize this with a lightweight head attached to a frozen estimator~\cite{qi2024hoisdf,wang2025unihope,zhou2026vpho}.
From the estimator's pixel-aligned features~\cite{saito2019pifu} and each vertex's distance to the predicted object, the head predicts the vertex's distance to the object in the lifted annotation.
We train one head per estimator on lifted annotations of a hand-object dataset.
Since the estimator itself is never updated, the head attaches to published estimators without retraining them.

%% file: sec/4_Evaluation.tex
\section{Evaluation}
\label{sec:evaluation}

\subsection{Experimental Setup}
\label{sec:eval_setup}

\paragraph{Datasets.}
We evaluate the lift on hand-object interaction datasets.
DexYCB~\cite{chao2021dexycb} is a multi-view RGB-D dataset of hands grasping YCB objects, and OakInk~\cite{yang2022oakink} is a large-scale collection of grasps on everyday objects.
ContactPose~\cite{brahmbhatt2020contactpose} records grasp contact as thermal imprints.

\paragraph{Implementation details.}
For estimated hands, we lift the outputs of three published estimators: HOISDF~\cite{qi2024hoisdf}, \mbox{UniHOPE}~\cite{wang2025unihope}, and VPHO~\cite{zhou2026vpho}.
We set \mbox{$\epsilon = 0.2$\,mm} and \mbox{$\tau = 1$\,mm}, and use $(w_{\mathrm{att}}, w_{\mathrm{rep}}, w_{\mathrm{prior}}) = (10^{4}, 500, 1)$ for seating, $(\lambda_{\mathrm{att}}, \lambda_{\mathrm{pen}}, \lambda_{\mathrm{reg}}, \lambda_{\mathrm{ver}}) = (10^{6}, 500, 100, 10^{6})$ for skin deformation, and $(\lambda_{\mathrm{vel}}, \lambda_{\mathrm{acc}}, \lambda_{\mathrm{wrist}}) = (10, 100, 10^{6})$ for sequences.
Each stage is solved with Adam~\cite{kingma2014adam}.
Further details are provided in \Cref{app:setup}, and the \href{https://shpark97.github.io/BeyondRigidHands}{code} will be released upon publication.

\paragraph{Baselines.}
We compare against two refiners that make rigid hands more plausible by changing the pose.
ContactOpt~\cite{grady2021contactopt} predicts target contact with a network and optimizes the hand pose of each frame toward it, and GeneOH Diffusion~\cite{liu2024geneoh} denoises hand-object motion with a diffusion model.

\vspace*{-0.3em}
\paragraph{Evaluation metrics.}
For contact, we report F1 against thermal maps~\cite{grady2021contactopt} and band conformity (C2) as the share of near-contact palmar skin on the surface.
For penetration, we report penetration depth (P.D.), intersection volume (I.V.)~\cite{hasson2019obman}, and the vertices over $1$\,mm deep ($N_{\mathrm{in}>1}$).
For stability, we report simulation displacement (S.D.)~\cite{hasson2019obman} and the force-equilibrium residual (F.E.)~\mbox{\cite{murray1994mathematical,hu2022physical,luo2024physics}} averaged over $G$ gravity directions (@$G$).
F.E. is the fraction of the object's weight that contacts cannot balance, and F.E.$_{\mathrm{fix}}$ fixes the contact count.
For pose, we report mean per-joint and per-vertex position errors (MPJPE, MPVPE) and acceleration error (Accel)~\cite{kocabas2020vibe}.
Full definitions are in \Cref{app:setup}.

\subsection{Refining Estimated Hands}
\label{sec:eval_posthoc}
\begin{figure}[t]
  \centering
  \includegraphics[width=\linewidth]{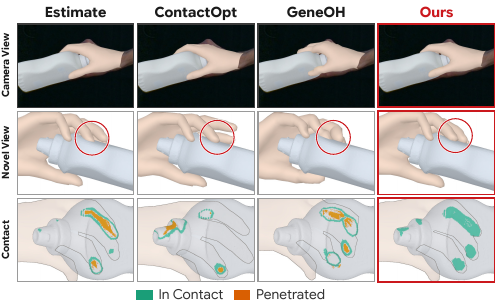}
  \caption{Refining a VPHO~\cite{zhou2026vpho} estimate.
  Rows show the camera view, a novel view, and contact on the object surface.
  Circles mark the finger pad.
  The estimate and the pose refiners~\cite{grady2021contactopt,liu2024geneoh} leave contact (\textcolor[HTML]{1A9E78}{green}) only as rims around penetrating regions (\textcolor[HTML]{D95F02}{orange}), whereas our lift forms conforming contact patches.}
  \label{fig:posthoc_clip}
\end{figure}
\addtocounter{figure}{1}
\begin{figure*}[t]
  \centering
  \includegraphics[width=\linewidth]{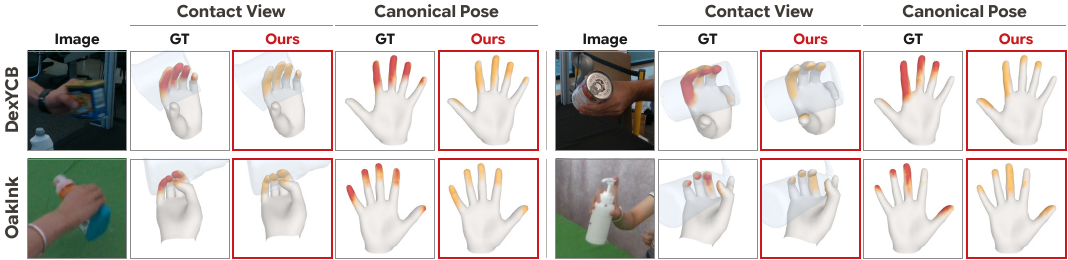}
  \caption{Rigid annotations (GT) and our lifts on DexYCB~\cite{chao2021dexycb} (top) and OakInk~\cite{yang2022oakink} (bottom), seen from the contact side and in the flat canonical pose.
  Colors show the signed distance to the object, with \textcolor[HTML]{8C8782}{grey} for free skin, \textcolor[HTML]{F28C28}{orange} for the contact band, and \textcolor[HTML]{B01818}{red} for skin inside the object.
  The annotation penetrates wherever it presses, whereas the lift forms wide surface patches at the annotated pose.}
  \label{fig:qual_pairs}
\end{figure*}

\paragraph{Contact at the estimated pose.}
To test whether contact can be added while keeping the estimated pose, we lift the outputs of the three estimators on the DexYCB dataset with geometry alone.
As shown in \Cref{tab:posthoc} and \Cref{fig:posthoc_clip}, the lift more than doubles band conformity for every estimator from about $20\%$ to nearly $50\%$ and removes most deeply penetrating vertices.
Meanwhile, the joint error grows by at most $0.3$\,mm and the acceleration error barely changes.
The pose refiners behave in the opposite way.
They lower band conformity below that of the estimate, add \mbox{$4$ to $20$\,mm} of joint error, and multiply the acceleration error several times.
They thus gain plausibility by moving the hand, whereas the lift creates contact in the skin and keeps the estimate nearly intact.
Geometry alone already preserves the stability of every estimate and improves it on HOISDF, yet it reaches only skin already near the object and leaves open the gaps that the learned head closes next.

\begin{table}[t]
  \centering
  \caption{Learning contact for estimated hands on the DexYCB~\cite{chao2021dexycb} dataset.
  F1 scores contact predicted from the estimated geometry and by the learned head against the lifted annotation.
  The other columns compare the lift without and with the head, where F1$_{\mathrm{GT}}$ scores the refined contact against the annotated one.}
  \label{tab:learned}
  \vspace{-0.75em}
  \small
  \resizebox{\linewidth}{!}{%
  \begin{tabular}{l|c|cccc}
    \toprule
    & \textbf{Geom.\,$\to$\,Head} & \multicolumn{4}{c}{\textbf{Lift\,$\to$\,Lift + Head}} \\
    \cmidrule(lr){2-2}\cmidrule(l){3-6}
    \textbf{Source} & F1 $\uparrow$ & F1$_{\mathrm{GT}}$ $\uparrow$ & C2 $\uparrow$ & S.D.@6 $\downarrow$ & F.E.@6 $\downarrow$ \\
    \midrule
    HOISDF~\cite{qi2024hoisdf} & 0.37\,$\to$\,0.57 & 0.31\,$\to$\,0.35 & 48.6\,$\to$\,51.7 & 2.19\,$\to$\,1.86 & 16.1\,$\to$\,11.1 \\
    UniHOPE~\cite{wang2025unihope} & 0.27\,$\to$\,0.46 & 0.17\,$\to$\,0.19 & 47.5\,$\to$\,49.6 & 5.27\,$\to$\,5.18 & 46.5\,$\to$\,44.4 \\
    VPHO~\cite{zhou2026vpho} & 0.47\,$\to$\,0.55 & 0.37\,$\to$\,0.39 & 46.7\,$\to$\,49.3 & 1.53\,$\to$\,1.39 & \phantom{0}9.3\,$\to$\,\phantom{0}7.0 \\
    \bottomrule
  \end{tabular}}
\end{table}
\addtocounter{figure}{-2}
\begin{figure}[t]
  \centering
  \includegraphics[width=\linewidth]{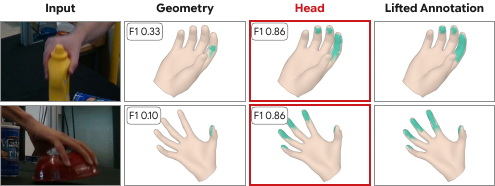}
  \caption{Contact predicted on two HOISDF~\cite{qi2024hoisdf} test frames.
  \textcolor[HTML]{1A9E78}{Green} marks hand vertices in contact from geometry, the learned head, and the lifted annotation that serves as the F1 reference.}
  \label{fig:learned}
\end{figure}
\addtocounter{figure}{1}
\begin{figure*}[t]
  \centering
  \includegraphics[width=\linewidth]{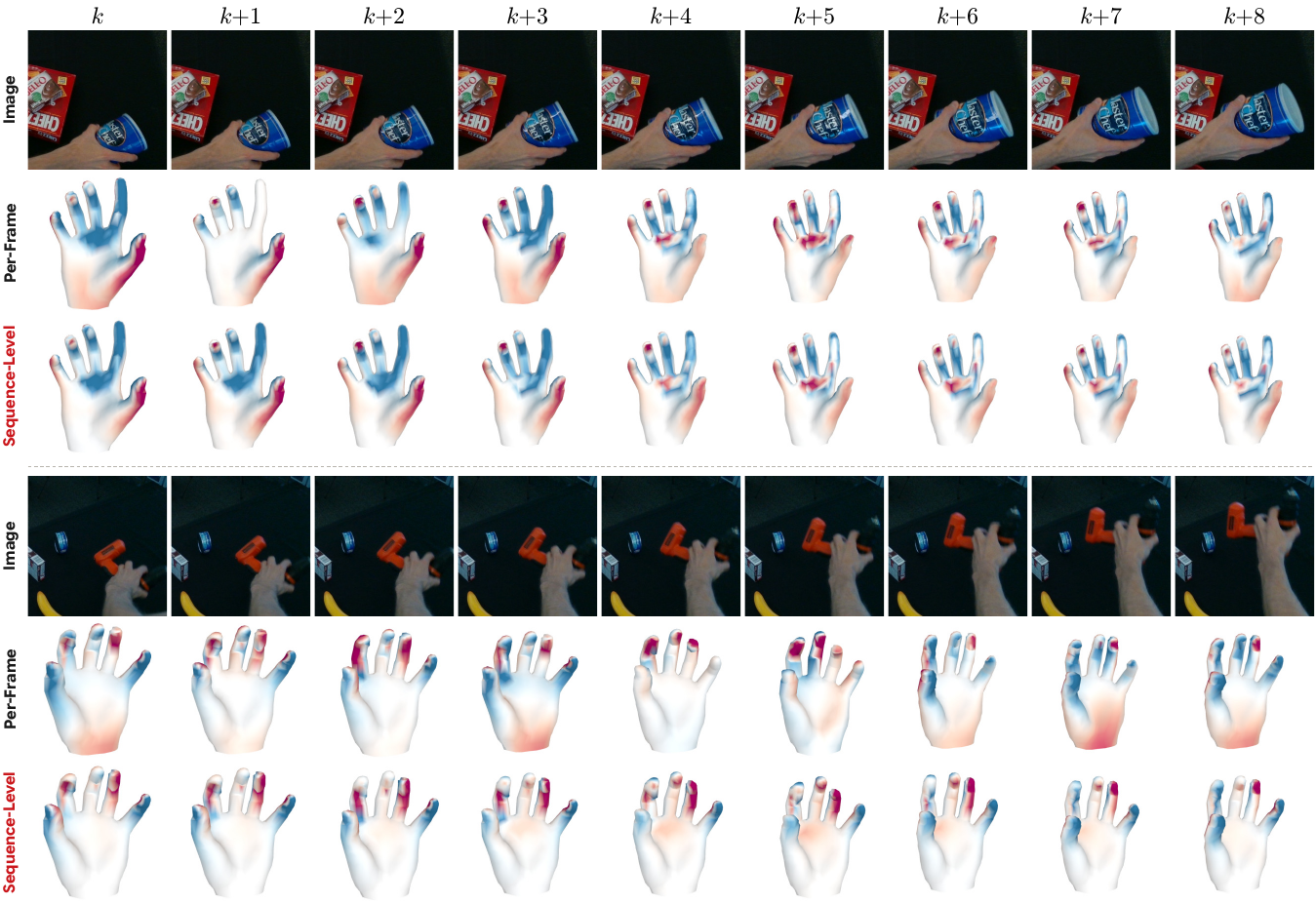}
  \caption{Per-frame and sequence-level lifts over nine consecutive frames of two DexYCB~\cite{chao2021dexycb} clips.
  Colors show the normal component of the canonical deformation, with \textcolor[HTML]{B2182B}{red} for inward and \textcolor[HTML]{2166AC}{blue} for outward displacement.
  The per-frame deformation flips sign on skin that stays in contact although the grasp barely changes, whereas the sequence-level lift evolves smoothly.}
  \label{fig:seq_fig}
\end{figure*}

\paragraph{Guidance from the learned head.}
To close the gaps that geometry misses, we guide skin deformation with the contact head of \Cref{sec:method_learn}.
As shown in \Cref{tab:learned} and \Cref{fig:learned}, the head predicts lifted contact more accurately than source geometry and raises F1 by $0.08$ to $0.20$ for every estimator.
With this guidance, the lift adds $2.1$ to $3.1$ points of band conformity, agrees better with the annotated contact, and lowers the equilibrium residual by $2.1$ to $5.0$ points on every estimator.
Among the refiners of \Cref{tab:posthoc}, the guided lift adds at most $0.4$\,mm of joint error and is the most stable on two of the three estimators with one tie in equilibrium residual.
On UniHOPE the predicted object is out of reach in $31\%$ of frames, and only ContactOpt is more stable at the cost of $15.6$\,mm of added joint error.
The learned contact thus turns into stability without giving up the estimated pose.

\subsection{Lifting GT Annotations}
\label{sec:dataset_results}

\paragraph{Contact and penetration.}
Like the estimates, the dataset annotations are rigid hands with gaps and penetrations.
We therefore lift $500$ test annotations each of DexYCB and OakInk, applying seating and skin deformation in turn to measure what each stage contributes.
As shown in \Cref{tab:dataset_plausibility}, seating alone cuts deeply penetrating vertices per frame from $38.6$ to $10.4$ on DexYCB and from $18.4$ to $8.0$ on OakInk.
Yet band conformity barely changes because the rigid surface still hovers above or penetrates the object around each contact.
Skin deformation then raises band conformity from about $25\%$ to $88\%$ on DexYCB and $71\%$ on OakInk, leaving fewer than three deeply penetrating vertices per frame.
\Cref{fig:qual_pairs} shows the lift forming wide surface patches where the rigid annotation penetrates.
Most of the gain thus comes from skin deformation, not the small pose correction.

\begin{table}[t]
  \centering
  \caption{Applying the lifting stages in turn to $500$ DexYCB and $500$ OakInk test annotations.
  Shaded rows give the full lift (ours).}
  \label{tab:dataset_plausibility}
  \vspace{-0.75em}
  \small
  \resizebox{\linewidth}{!}{%
  \begin{tabular}{l|cc|c|cc|cccc}
    \toprule
    & \multicolumn{2}{c|}{} & \textbf{Contact} & \multicolumn{2}{c|}{\textbf{Penetration}} & \multicolumn{4}{c}{\textbf{Stability}} \\
    \cmidrule(lr){4-4}\cmidrule(lr){5-6}\cmidrule(l){7-10}
    & &
      & \textbf{C2}
      & \textbf{P.D.} & \textbf{$N_{\mathrm{in}>1}$}
      & \textbf{S.D.@1} & \textbf{S.D.@6} & \textbf{F.E.@6} & \textbf{F.E.}$_{\mathrm{fix}}$ \\
    \textbf{Dataset} & \textbf{Seating} & \textbf{Skin}
      & {\small (\%) $\uparrow$}
      & {\small (mm) $\downarrow$} & {\small (\#) $\downarrow$}
      & {\small (cm) $\downarrow$} & {\small (cm) $\downarrow$} & {\small (\%) $\downarrow$} & {\small (\%) $\downarrow$} \\
    \midrule
      \multirow{3}{*}{DexYCB~\cite{chao2021dexycb}} &  &  & 19.4 & 5.97 & 38.6 & 1.85 & 1.67 & 11.6 & 9.2 \\
       & \cmark &  & \underline{24.9} & \underline{4.68} & \underline{10.4} & \underline{1.82} & \underline{1.56} & \underline{11.3} & \underline{8.9} \\
       & \gc\cmark & \gc\cmark & \gc \textbf{87.9} & \gc \textbf{4.48} & \gc \textbf{2.9} & \gc \textbf{1.64} & \gc \textbf{1.38} & \gc \textbf{7.9} & \gc \textbf{6.0} \\
    \midrule
      \multirow{3}{*}{OakInk~\cite{yang2022oakink}} &  &  & 19.7 & 5.18 & 18.4 & 1.90 & 1.49 & 13.1 & 11.3 \\
       & \cmark &  & \underline{23.8} & \textbf{4.62} & \underline{8.0} & \underline{1.89} & \underline{1.47} & \underline{10.9} & \underline{9.8} \\
       & \gc\cmark & \gc\cmark & \gc \textbf{71.4} & \gc \textbf{4.62} & \gc \textbf{1.5} & \gc \textbf{1.66} & \gc \textbf{1.33} & \gc \textbf{7.6} & \gc \textbf{7.6} \\
    \bottomrule
  \end{tabular}}
\end{table}

\paragraph{Grasp stability.}
We next ask whether the added contact makes the grasps physically more plausible, namely whether the contacts can better hold the object against gravity.
As shown in \Cref{tab:dataset_plausibility}, lifting reduces slip in simulation and lowers the equilibrium residual from $11.6\%$ to $7.9\%$ on DexYCB and from $13.1\%$ to $7.6\%$ on OakInk.
Seating alone does not significantly change slip on either dataset or equilibrium on DexYCB, so most of this gain comes from skin deformation.
The gain also persists at a fixed contact count, indicating that it stems from where the skin touches rather than from how much of it does.
Results on HO3D~\cite{hampali2020honnotate} follow the same trends (see \Cref{app:results}).

\begin{table}[t]
  \centering
  \caption{Agreement with thermal contact on ContactPose~\cite{brahmbhatt2020contactpose}.
  $\Delta$P.\ is the pose change (mean vertex displacement from the MANO fit), and the matched settings align the contact size or $\Delta$P.\ with ours.}
  \label{tab:contactpose}
  \vspace{-0.75em}
  \small
  \resizebox{\linewidth}{!}{%
  \begin{tabular}{l|c|c|ccc}
    \toprule
    \textbf{Method} & \textbf{Setting} & $\Delta$\textbf{P.} {\small (mm)} & \textbf{Prec.} $\uparrow$ & \textbf{Rec.} $\uparrow$ & \textbf{F1} $\uparrow$ \\
    \midrule
    \multirow{3}{*}{MANO Fit~\cite{brahmbhatt2020contactpose}}
      & Rigid Annotation & --- & \underline{0.72} & 0.33 & 0.43 \\
      & + Matched Size & --- & 0.70 & \underline{0.46} & \underline{0.54} \\
      & \gc Lift (ours) & \gc 2.9 & \gc \textbf{0.74} & \gc \textbf{0.49} & \gc \textbf{0.57} \\
    \midrule
    \multirow{3}{*}{ContactOpt~\cite{grady2021contactopt}}
      & Released Setting & 7.8 & 0.62 & 0.42 & 0.49 \\
      & Matched Pose Change & 3.6 & 0.71 & 0.40 & 0.49 \\
      & + Matched Size & 3.6 & 0.70 & \underline{0.46} & \underline{0.54} \\
    \bottomrule
  \end{tabular}}
\end{table}

\begin{table}[t]
  \centering
  \caption{The temporal prior against post-hoc smoothing on held-out clips.
  Filtering per-frame lifts raises transfer (Tr.)\ but lowers band conformity (C2), and re-projecting the filtered result restores conformity at the cost of transfer.
  Only the temporal prior keeps both high, with the highest harmonic mean (HM) of the two.}
  \label{tab:heldout_body}
  \vspace{-0.75em}
  \small
  \resizebox{\linewidth}{!}{%
  \begin{tabular}{l|ccc|ccc}
    \toprule
    & \multicolumn{3}{c|}{\textbf{DexYCB}} & \multicolumn{3}{c}{\textbf{OakInk}} \\
    \cmidrule(lr){2-4}\cmidrule(l){5-7}
    & \textbf{Tr.} & \textbf{C2} & \textbf{HM} & \textbf{Tr.} & \textbf{C2} & \textbf{HM} \\
    \textbf{Method} & {\small (\%) $\uparrow$} & {\small (\%) $\uparrow$} & {\small (\%) $\uparrow$} & {\small (\%) $\uparrow$} & {\small (\%) $\uparrow$} & {\small (\%) $\uparrow$} \\
    \midrule
    Per-Frame & 39.8 & \underline{87.5} & 54.7 & 34.5 & \underline{75.1} & 47.3 \\
    Post-hoc Filter & \underline{78.1} & 75.5 & \underline{76.8} & \underline{77.9} & 63.9 & \underline{70.2} \\
    Filter + Re-Projection & 65.2 & \textbf{89.4} & 75.4 & 58.7 & \textbf{76.9} & 66.6 \\
    \rowcolor{tabgray} Temporal Prior (ours) & \textbf{83.2} & 83.5 & \textbf{83.4} & \textbf{82.7} & 69.1 & \textbf{75.3} \\
    \bottomrule
  \end{tabular}}
\end{table}

\paragraph{Agreement with thermal contact.}
\label{sec:eval_contactpose}
ContactPose records grasp contact as thermal imprints on the object, a reference independent of any hand mesh.
We lift its $1{,}766$ grasps and follow ContactOpt~\cite{grady2021contactopt} in comparing object vertices within $2$\,mm of the hand with the thermal contact.
We also match the contact size to ours by enlarging the compared contact to as many object vertices, and the pose change by limiting ContactOpt to move the hand about as far as the lift.
As shown in \Cref{tab:contactpose}, lifting raises F1 from $0.43$ to $0.57$ mainly by recovering missed contact (recall $0.33$ to $0.49$) without losing precision.
The rigid contact reaches only $0.54$ at the matched size, so contact area alone does not fully explain the gain.
ContactOpt reaches no more, even at the matched pose change and size.
The lifted contact thus matches measured contact better than the rigid annotation and pose refinement.
Lifted annotations are also better supervision, since a contact estimator trained on them instead of rigid ones predicts thermal contact more accurately (see \Cref{app:results}).

\begin{figure}[!t]
  \centering
  \includegraphics[width=\linewidth]{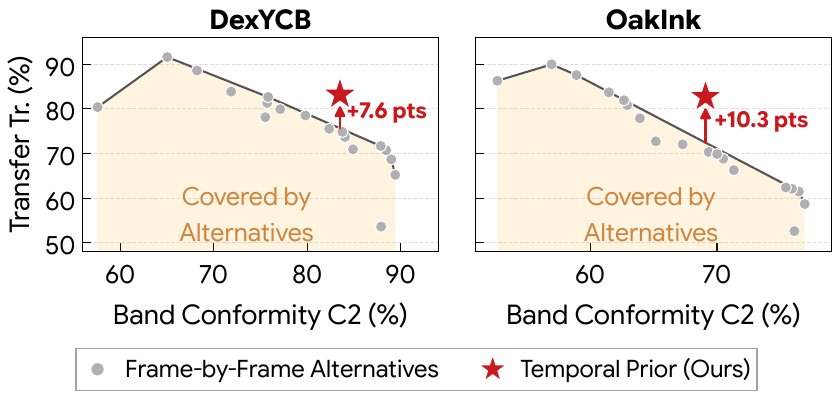}
  \caption{Transfer against band conformity on held-out clips.
  Each grey dot is one setting of filtering, re-projection, or warm-started tracking, and the shaded region lies below their upper envelope.
  The temporal prior (\textcolor[HTML]{CB181D}{star}) exceeds it by the marked margin.}
  \label{fig:app_frontier}
\end{figure}

\paragraph{Temporal consistency.}
\label{sec:eval_consistency}
Finally, we test on the DexYCB and OakInk datasets whether the temporal prior keeps the deformation consistent over whole clips.
We measure band conformity (C2) and transfer (Tr.), the share of canonical deformation that persists between adjacent frames.
We compare against per-frame lifts, their temporal filtering, and filtering followed by per-frame re-optimization (re-projection).
\mbox{\Cref{fig:seq_fig}} shows per-frame lifts flipping the deformation sign on skin that stays in contact.
As shown in \Cref{tab:heldout_body}, the prior raises transfer from $39.8\%$ to $83.2\%$ on DexYCB and from $34.5\%$ to $82.7\%$ on OakInk while staying within $4$ and $6$ points of the per-frame conformity.
Filtering approaches this transfer but loses $12$ and $11$ points of conformity, and re-projection restores conformity at the cost of transfer.
The prior has the highest harmonic mean (HM) of the two on both datasets.
\Cref{fig:app_frontier} extends this comparison to $19$ settings of filtering, re-projection, and warm-started tracking.
At equal conformity, even their upper envelope falls $7.6$ and $10.3$ points short of the prior in transfer.
Transfer and conformity thus come together only with joint optimization.

\subsection{Further Analysis}
\label{sec:eval_ablation}

\begin{table}[t]
  \centering
  \caption{Simpler alternatives to the lift.
  Stretch, Flips, and Change measure skin distortion and its change between frames.
  All columns average the three estimators of \Cref{tab:posthoc} except Depth, the residual to observed depth on DexYCB annotations.
  $\dagger$ adds extra smoothing.}
  \label{tab:alternatives}
  \vspace{-0.75em}
  \small
  \resizebox{\linewidth}{!}{%
  \begin{tabular}{l|cc|ccc|c}
    \toprule
    & \textbf{C2} & \textbf{F.E.@6} & \textbf{Stretch} & \textbf{Flips} & \textbf{Change} & \textbf{Depth} \\
    \textbf{Method} & {\small (\%) $\uparrow$} & {\small (\%) $\downarrow$} & {\small (\%) $\downarrow$} & {\small (\#) $\downarrow$} & {\small (mm) $\downarrow$} & {\small (mm) $\downarrow$} \\
    \midrule
    Rigid Input & 21.0 & 24.5 & --- & --- & --- & 4.75 \\
    Pose Space & 20.1 & 26.3 & \textbf{9} & \textbf{0.02} & \underline{1.15} & \underline{5.21} \\
    Contact-Band Rule & \textbf{51.5} & \underline{24.2} & 40 & 1.86 & 2.86 & 6.10 \\
    Contact-Band Rule$^{\dagger}$ & 29.1 & 26.0 & \underline{24} & 0.79 & 2.43 & --- \\
    \rowcolor{tabgray} Lift (ours) & \underline{47.6} & \textbf{24.0} & 25 & \underline{0.09} & \textbf{1.04} & \textbf{4.91} \\
    \bottomrule
  \end{tabular}}
\end{table}

\paragraph{Simpler alternatives.}
We first ask whether a simpler correction could replace the lift.
A pose-space variant keeps every skinning segment rigid, while a closed-form contact-band rule~\cite{sorkine2004laplacian} projects the contact band onto the object and smooths the rest.
Since contact can come from distorting the skin, we also measure the distortion: Stretch as the change in edge length, Flips as the number of inverted faces, and Change as the difference in deformation between frames.
As shown in \Cref{tab:alternatives}, the pose-space variant cannot bend the surface onto the object and leaves conformity at the input level and the grasps less stable.
The rule conforms slightly more than the lift, but with $1.6$ times the stretch, about $20$ times the flips, and almost three times the change.
With extra smoothing that matches the stretch of the lift, the rule conforms only $29\%$ of the band against $48\%$.
Only the lift thus conforms the skin without distorting it.

\paragraph{Agreement with observed depth.}
We next ask whether creating contact by moving the skin takes it away from the real hand.
We test this with DexYCB depth images, which record the real surface and are not used by the lift.
We measure Depth as the difference between the rendered depth of each hand and the captured depth on the visible skin next to the object.
As shown in \Cref{tab:alternatives}, the lift keeps Depth at $4.91$ against $4.75$ for the rigid input even though that annotation was fitted to these images.
The pose-space variant reaches $5.21$ without gaining contact, and the contact-band rule reaches $6.10$ for a similar conformity.
The lift thus adds contact without leaving the observed surface.

\begin{table}[t]
  \centering
  \caption{Source of the head's gain.
  Each entry is the residual of the guided lift minus that of a variant (negative favors the head).
  Nearest Vertices adds as many vertices as the head by proximity to the object, and Rigid-Label Head is trained on rigid annotations.}
  \label{tab:head_controls}
  \vspace{-0.75em}
  \small
  \resizebox{\linewidth}{!}{%
  \begin{tabular}{l|ccc|ccc}
    \toprule
    & \multicolumn{3}{c|}{$\Delta$\textbf{F.E.@6} {\small (\%) $\downarrow$}} & \multicolumn{3}{c}{$\Delta$\textbf{F.E.}$_{\mathrm{fix}}$ {\small (\%) $\downarrow$}} \\
    \cmidrule(lr){2-4}\cmidrule(l){5-7}
    \textbf{Compared with} & HOISDF~\cite{qi2024hoisdf} & UniHOPE~\cite{wang2025unihope} & VPHO~\cite{zhou2026vpho} & HOISDF~\cite{qi2024hoisdf} & UniHOPE~\cite{wang2025unihope} & VPHO~\cite{zhou2026vpho} \\
    \midrule
    Lift & $-4.98$ & $-2.11$ & $-2.23$ & $-4.43$ & $-2.78$ & $-2.18$ \\
    Lift + Nearest Vertices & $-0.98$ & $-0.45$ & $-0.44$ & $-1.00$ & $-0.74$ & $-0.44$ \\
    Lift + Rigid-Label Head & $-1.60$ & $-1.03$ & $-0.84$ & $-1.38$ & $-1.69$ & $-0.81$ \\
    \bottomrule
  \end{tabular}}
\end{table}

\paragraph{Source of the head's gain.}
The learned head decides where the skin of an estimated hand should touch, so we test with two controls whether its stability gain comes from the lifted labels or merely from adding contact.
Nearest vertices adds as many attraction vertices as the head but takes those closest to the object, while the rigid-label head guides the lift with the same head trained on rigid annotations.
Since more contact alone lowers the equilibrium residual, we also compare at an equal contact count (F.E.$_{\mathrm{fix}}$).
As shown in \Cref{tab:head_controls}, the head lowers the residual of the unguided lift by $2.1$ to $5.0$ points and keeps this gain at an equal contact count.
Nearest vertices leaves the residual $0.4$ to $1.0$ points higher and the rigid-label head $0.8$ to $1.7$ points higher on every estimator under both measures.
The gain thus comes from where the head adds contact rather than how much, and it learns this placement only from lifted labels.

%% file: sec/5_Conclusion.tex
\section{Conclusion}
\label{sec:conclusion}

Whether annotated or estimated, rigid hand meshes hover above or sink into the objects they grasp.
We lift such hands to contact-consistent meshes by seating them on the object, deforming the skin where it meets the surface, and keeping this deformation coherent over time.
Unlike pose refiners, the lift increases skin contact on three published estimators while leaving the pose nearly unchanged.
A head learned from lifted annotations further stabilizes the grasps.
On dataset annotations, the lift removes most deep penetration, agrees better with thermal contact, stays faithful to captured depth, and yields labels that train better contact estimators.
Since the lifted mesh keeps the MANO topology and the head attaches to frozen estimators, existing datasets and pipelines can adopt contact-consistent hands as they are.

\section*{LLM Usage Statement}
We used Claude (Anthropic) to assist with writing code, running experiments, creating figures, editing the text for readability, and formatting references.
The authors directed the research, reviewed all AI-assisted work, and verified the code and results, taking full responsibility for the final content, including text, claims, and artifacts.

%% file: sec/6_Supplementary.tex
\onecolumn
\begin{center}
  {\Large\textbf{\thetitle}\\[0.5em]Supplementary Material}
\end{center}
\vspace{1.0em}
\raggedbottom
\makeatletter\setlength{\@fptop}{0pt}\makeatother

\section{Additional Qualitative Results}
\label{app:qual}
This section provides additional qualitative results of our lift that extend the comparisons in the main paper.
We first visualize the contact that the learned head predicts on estimated hands in \Cref{app:qual_head}.
We then compare further rigid annotations with their lifts in \Cref{app:qual_annot}.
Finally, we contrast per-frame and sequence-level lifts on two clips in \Cref{app:qual_seq}.

\subsection{Learned Contact on Estimated Hands}
\label{app:qual_head}
\begin{figure}[H]
  \centering
  \includegraphics[width=\textwidth]{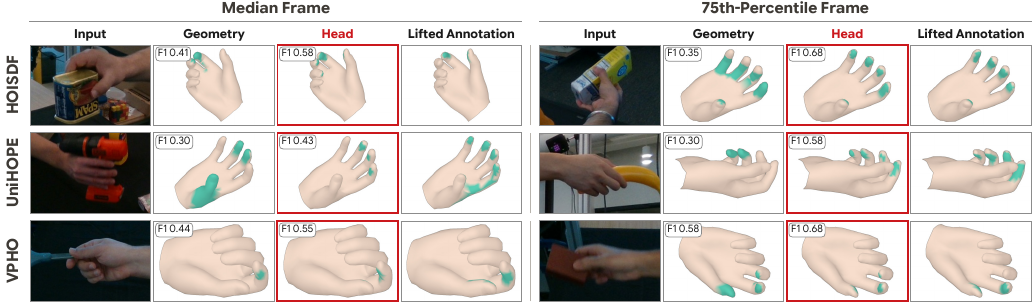}
  \caption{Contact from geometry, the learned head, and the lifted annotation on DexYCB frames at the median (left) and the $75$th percentile (right) of the head's F1. \textcolor[HTML]{1A9E78}{Green} marks vertices in contact, with F1 against the lifted annotation in each panel.}
  \label{fig:app_learned_rank}
\end{figure}

\Cref{fig:app_learned_rank} compares the contact predicted by the learned head with that of the estimated geometry on additional DexYCB frames.
For each estimator we rank the frames by the head's F1 against the lifted annotation and show the median as a typical case and the $75$th percentile as a favorable one.
The head matches the lifted contact more closely at the $75$th percentile than at the median, yet it remains more accurate than geometry in both cases.
It concentrates contact on the fingertips that touch the object, whereas geometry misses part of this contact or spreads it over skin that does not touch.

\subsection{Lifted GT Annotations}
\label{app:qual_annot}
\begin{figure}[H]
  \centering
  \includegraphics[width=\textwidth]{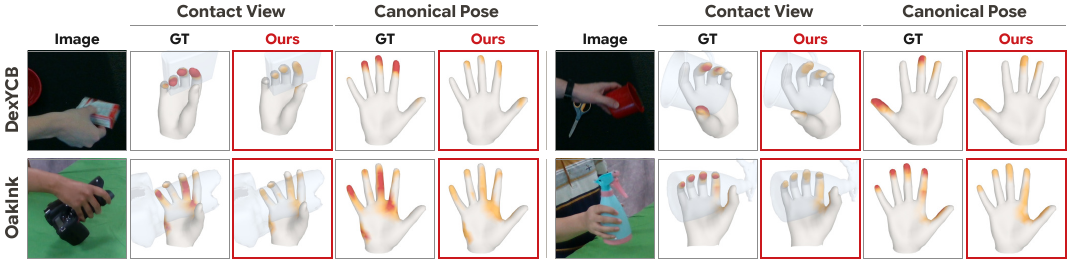}
  \caption{Rigid annotations (GT) and our lifts on DexYCB and OakInk frames, seen from the contact side and in the flat canonical pose. Skin is colored by its signed distance to the object as free (\textcolor[HTML]{8C8782}{grey}), in the contact band (\textcolor[HTML]{F28C28}{orange}), or inside the object (\textcolor[HTML]{B01818}{red}).}
  \label{fig:qual_pairs_more}
\end{figure}

\Cref{fig:qual_pairs_more} compares rigid annotations with our lifts on additional DexYCB and OakInk frames, seen from the contact side and in the flat canonical pose that shows where on the hand the contact lies.
The rigid annotations sink into the object where the fingers press, leaving only thin rims of skin in the contact band.
At the same pose, our lifts flatten this skin onto the surface and turn the rims into broad contact patches on the finger pads.

\subsection{Temporal Consistency}
\label{app:qual_seq}
\begin{figure}[H]
  \centering
  \includegraphics[width=0.95\linewidth]{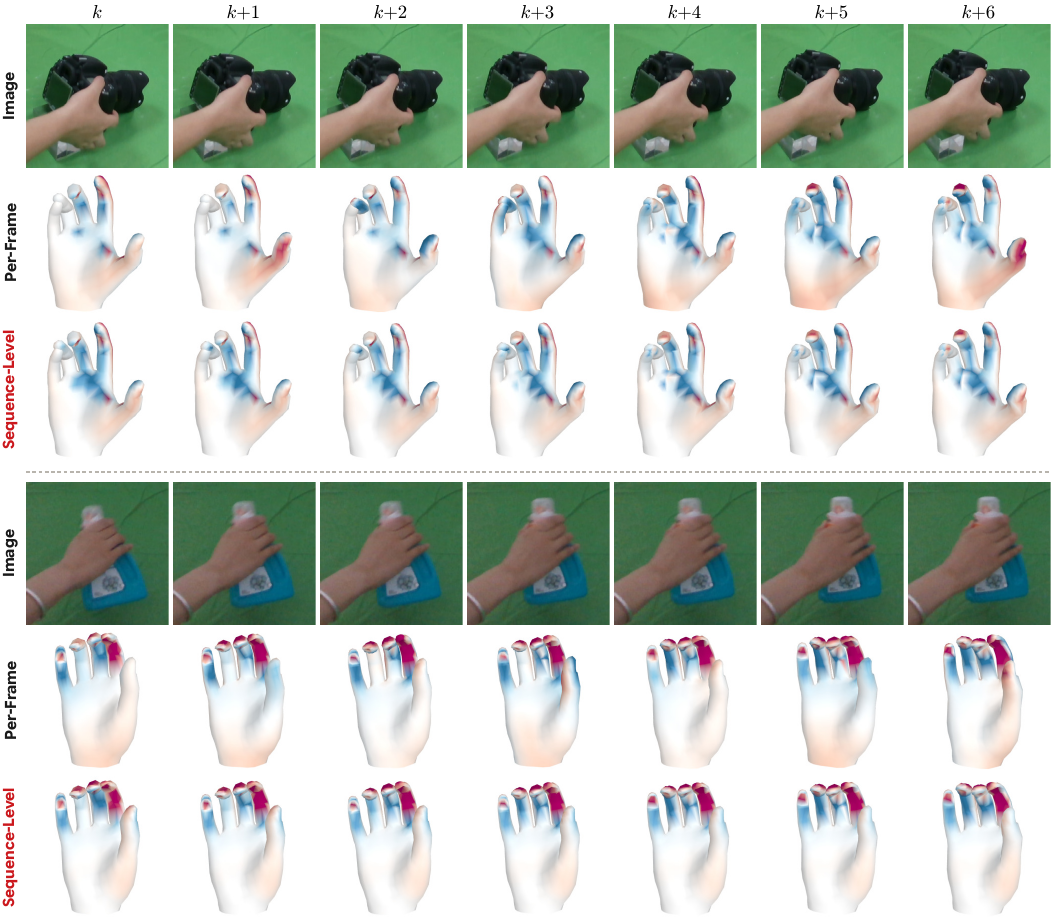}
  \caption{Per-frame and sequence-level lifts on seven consecutive frames of two OakInk clips with a camera (top) and a bottle (bottom). Skin is colored by the normal component of its canonical deformation as inward (\textcolor[HTML]{B2182B}{red}) or outward (\textcolor[HTML]{2166AC}{blue}) displacement.}
  \label{fig:seq_fig_abl}
\end{figure}

\Cref{fig:seq_fig_abl} compares per-frame and sequence-level lifts on two OakInk clips, colored by the normal component of the canonical deformation.
Since this deformation is expressed as if the hand were at rest, a change in color reflects a change of the skin rather than motion of the hand.
Per-frame lifting leaves the skin free to slide along the object, so the indentation flickers between neighboring frames although the grasp barely changes.
The sequence-level lift solves all frames jointly and keeps the indentation steady over the clip.

\section{Experimental Details}
\label{app:setup}
This section extends the experimental setup of the main paper with additional details of our lift.
We first describe the implementation details of the lift and the learned head in \Cref{app:impl}.
We then present the loss terms in \Cref{app:losses}.
Next, we summarize the datasets with their splits, filters, and preprocessing in \Cref{app:data}.
Finally, we detail the evaluation protocols and metrics in \Cref{app:protocol}.

\subsection{Implementation Details}
\label{app:impl}
Seating minimizes Eqn.~\eqref{eq:kin_obj} with Adam~\cite{kingma2014adam} for $120$ steps and projects the correction after each step onto the box $\mathcal{C}$, which allows only a slight bend of each joint and a small shift of the wrist.
Skin deformation then minimizes Eqn.~\eqref{eq:dataset_loss} from $\mathbf{u} = \mathbf{0}$ for $1{,}500$ iterations with a step size of $0.1$ that decays during optimization.
The sequence-level lift solves Eqn.~\eqref{eq:seq_obj} jointly over clips of up to $45$ frames.
Because predicted objects jitter between frames, the temporal prior also couples the seating variables of estimated inputs.
None of the lift's weights is tuned on test data.
On annotations, seating shifts the wrist by $1$ to $2$\,mm and skin deformation then moves the vertices by about $1$\,mm.

The learned head predicts the signed distance $\hat{s}_i$ of each hand vertex $i$ to the object in the lifted annotation.
It samples the pixel-aligned features~\cite{saito2019pifu} of the frozen estimator at the projected vertices, reduces them to $128$ channels by PCA, and combines them with the signed distance to the predicted object and a vertex embedding in a per-vertex MLP.
Two graph layers~\cite{hamilton2017graphsage} on the MANO mesh~\cite{romero2017mano} then output $\hat{s}_i$ with a Laplace scale as its uncertainty~\cite{kendall2017uncertainties}.
We train one head per estimator on the lifted DexYCB s0 training annotations with a negative log-likelihood loss and a smoothness hinge along mesh edges, selecting the checkpoint on the validation split.
Since the estimators fit their own training images almost exactly, we query the head at annotations perturbed in hand pose and object placement to mimic the errors of estimates.
At test time, vertices within $15$\,mm of the object with $\hat{s}_i \leq 2$\,mm join the attraction set $\mathcal{B}_{\mathrm{a}}$ of Eqn.~\eqref{eq:app_attach}, while the contact band and seating stay unchanged.
This rule is tuned on $20$ of the $80$ test clips, and the other $60$ serve as held-out clips in \Cref{app:res_posthoc}.

\subsection{Loss Terms}
\label{app:losses}
We define the terms of Eqns.~\eqref{eq:kin_obj}, \eqref{eq:dataset_loss}, and~\eqref{eq:seq_obj} on the seated hand $\mathcal{V}'$ and the lifted hand $\Phi = \mathcal{V}' + \mathbf{u}$.
The contact terms use the signed distance $\Omega$ to the object surface~\cite{park2019deepsdf} and the penetration depth $\phi = [-\Omega]_{+}$ into the object.

\paragraph{Kinematic seating.}
Following ContactOpt~\cite{grady2021contactopt}, seating pulls the palmar skin near the object to a target depth $\delta_{\mathrm{skin}} = 2$\,mm inside the surface while penalizing deeper penetration and large pose corrections,
\begin{align}
  \mathcal{L}'_{\mathrm{att}} &= \frac{1}{|\mathcal{B}'|}\sum_{i \in \mathcal{B}'} \big(\Omega(\mathcal{V}'_i) + \delta_{\mathrm{skin}}\big)^2, \label{eq:app_seat_att}\\
  \mathcal{L}'_{\mathrm{rep}} &= \frac{1}{|\mathcal{P}'|}\sum_{i \in \mathcal{P}'} e'_i + \max_i\, e'_i, \qquad e'_i = \big[\phi(\mathcal{V}'_i) - \delta_{\mathrm{skin}}\big]_{+}, \label{eq:app_seat_rep}\\
  \mathcal{L}_{\mathrm{prior}} &= \frac{\|\alpha\|^2}{15\,\sigma_{\alpha}^2} + \frac{\|\Delta\mathbf{t}\|^2}{\sigma_{\mathrm{t}}^2}, \label{eq:app_seat_prior}
\end{align}
where $\mathcal{B}'$ holds the palmar vertices from $\delta_{\mathrm{skin}}$ inside to $\tau_{\mathrm{c}} = 5$\,mm outside the surface and $\mathcal{P}'$ holds those deeper than $\delta_{\mathrm{skin}}$.
The prior normalizes $\alpha$ and $\Delta\mathbf{t}$ by $\sigma_{\alpha} = 0.1$ and $\sigma_{\mathrm{t}} = 2$\,mm, and the box $\mathcal{C}$ limits them to $\alpha_j \in [-0.10, 0.15]$ and $\|\Delta\mathbf{t}\|_{\infty} \leq 8$\,mm.
Seating thus leaves a shallow penetration of $\delta_{\mathrm{skin}}$, which skin deformation then flattens onto the surface.

\paragraph{Skin deformation.}
This stage instead draws the skin onto the surface itself through the terms of Eqn.~\eqref{eq:dataset_loss},
\begin{align}
  \mathcal{L}_{\mathrm{pen}} &= \frac{1}{|\mathcal{P}|}\sum_{i \in \mathcal{P}} e_i + \max_i\, e_i, \qquad e_i = \big[\phi(\Phi_i) - \epsilon_i\big]_{+}, \label{eq:app_pen}\\
  \mathcal{L}_{\mathrm{att}} &= \frac{1}{|\mathcal{B}_{\mathrm{a}}|}\sum_{i \in \mathcal{B}_{\mathrm{a}}} \eta_i\, \Omega(\Phi_i)^2, \label{eq:app_attach}\\
  \mathcal{L}_{\mathrm{ver}} &= \frac{1}{N}\sum_{i=1}^{N} \big[ \|\mathbf{u}_i\| - b_i \big]_{+}^2, \label{eq:app_umax}
\end{align}
where $\mathcal{P}$ holds the penetrating vertices and the max term keeps any single vertex from staying deep.
The allowance $\epsilon_i$ equals $\epsilon$ except on the vertices $\mathcal{X}$ exempted from C1 that the source places beyond the trust band $\tau_{\mathrm{t}} = 5$\,mm, where it keeps the source depth.
The attraction set $\mathcal{B}_{\mathrm{a}}$ holds the vertices within $5$\,mm of the object at the seated pose.
Their weight $\eta_i = \rho_i + 0.1\,(1 - \rho_i)$ follows a palmar score $\rho_i$, so dorsal skin is drawn only weakly.
The cap $b_i$ softly bounds each displacement at the seated penetration depth plus $5$\,mm.
In Eqn.~\eqref{eq:reg}, we compute $\nabla_{\!f}\, \mathbf{u}$ as in Aigerman~\etal~\cite{NJF} and confine deformation to the skin near contact by tripling the stiffness $m_f$ on dorsal faces and reducing it fivefold on palmar faces within $10$\,mm of the object.

\paragraph{Sequence-level optimization.}
A frame enters the pairs $\mathcal{K}_1$ and triples $\mathcal{K}_2$ of Eqn.~\eqref{eq:temporal} only when its attraction set is nonempty, which keeps frames before contact undeformed.
Because the open wrist boundary of the mesh would otherwise drift to lower $\mathcal{L}_{\mathrm{reg}}$, each frame anchors the vertices $\mathcal{W}$ within $35$\,mm of the wrist and outside the contact band by
\begin{equation}
  \label{eq:wrist}
  \mathcal{L}_{\mathrm{wrist}} = \frac{1}{|\mathcal{W}|} \sum_{i \in \mathcal{W}} \|\mathbf{u}_i\|^2.
\end{equation}

\subsection{Data}
\label{app:data}
We lift DexYCB~\cite{chao2021dexycb} and OakInk~\cite{yang2022oakink} frames in which the hand grasps the object and both are visible, using the official s0 split for DexYCB.
A closure filter keeps frames with at least four hand vertices within $5$\,mm of the object and contacts near force closure~\cite{liu2021synthesizing}, removing frames where the hand approaches, releases, or unstably holds the object.
Visibility filters further require at least four of the $21$ joints and $90\%$ of the object vertices to project into the image.
As the retained grasps are already near force closure, our stability gains are conservative.
For each object, we precompute signed-distance and penetration-depth grids after making the templates watertight~\cite{huang2020manifoldplus}.
We also recover MANO parameters for each annotated hand mesh.
On HO3Dv2~\cite{hampali2020honnotate} we fit MANO to the released hand meshes and lift clips with seating coupled over time, while on ContactPose~\cite{brahmbhatt2020contactpose} we lift the released MANO fits.
We recover its thermal maps by inverting the colormap of the contact textures in the official ContactPose web viewer.
These match the official maps almost exactly where both exist.

\subsection{Evaluation Protocols and Metrics}
\label{app:protocol}
We evaluate refinement on DexYCB clips unseen by the estimators.
Each clip holds the first $45$ frames of the longest run in a right-hand s0 test sequence where the annotated hand stays within $3$\,mm of the object.
HOISDF~\cite{qi2024hoisdf}, UniHOPE~\cite{wang2025unihope}, and VPHO~\cite{zhou2026vpho} run their released checkpoints with their own predicted objects and place the hand at the annotated wrist as in their protocols.
ContactOpt~\cite{grady2021contactopt} runs its released per-frame refinement and a Matched Pose Change variant whose lower learning rate moves the vertices about as far as the lift.
GeneOH Diffusion~\cite{liu2024geneoh} runs its released denoiser with the final MANO fit initialized at the source pose.
Every refiner receives only the source hand and object and is scored against the same predicted object.
All comparisons report paired $95\%$ bootstrap intervals clustered by clip or sequence.

\paragraph{Contact and penetration.}
Band conformity is the share of the contact band $\mathcal{B}$ that lies on the object surface,
\begin{equation}
  \label{eq:app_c2}
  \mathrm{C2} = \frac{1}{|\mathcal{B}|} \sum_{i \in \mathcal{B}} \mathds{1}\big[\,|\Omega(\Phi_i)| \leq \tau\,\big],
\end{equation}
where $\mathds{1}[\cdot]$ is the indicator function and $\tau = 1$\,mm.
The band $\mathcal{B}$ holds the palmar vertices within $\tau_{\mathrm{c}}$ of the object at the seated pose and comes from the source hand for estimates.
P.D.\ and I.V.~\cite{hasson2019obman} are computed from object-surface voxels inside the hand.
Because voxelization limits the resolution of P.D., $N_{\mathrm{in}>1}$ counts the hand vertices deeper than $1$\,mm including those exempted from C1.
Distances and inside tests use dense surface samples and winding numbers~\cite{jacobson2013winding} rather than the optimization grids.

\paragraph{Stability.}
S.D.~\cite{hasson2019obman,wang2023dexgraspnet} holds the hand fixed, simulates the object under gravity, and reports how far its center of mass moves.
S.D.@1 applies gravity downward and S.D.@6 averages over six directions.
F.E.\ is the fraction of the object weight that the contacts cannot balance~\cite{murray1994mathematical,hu2022physical,luo2024physics},
\begin{equation}
  \label{eq:app_fe}
  \mathrm{F.E.} = \min_{\mathbf{f} \geq \mathbf{0}} \big\| Q \mathbf{f} - \mathbf{w} \big\|_2,
\end{equation}
where $\mathbf{w}$ is a unit gravity wrench and $Q$ stacks friction-cone wrenches ($\mu = 0.5$) of hand vertices within $3$\,mm of the surface.
Frames with under three contacts count as fully unbalanced, and F.E.@6 averages over six directions.
F.E.$_{\mathrm{fix}}$ equalizes contact counts between compared hands by farthest-point sampling~\cite{qi2017pointnetpp}, so it reflects where the skin touches rather than how much.

\paragraph{Pose and temporal consistency.}
MPJPE averages the joint error over the $21$ hand joints $\mathcal{J}$ after aligning the wrists,
\begin{equation}
  \label{eq:app_mpjpe}
  \mathrm{MPJPE} = \frac{1}{|\mathcal{J}|}\sum_{j \in \mathcal{J}}\vphantom{\sum_{j=0}^{20}} \big\| (\mathbf{J}_j - \mathbf{J}_0) - (\bar{\mathbf{J}}_j - \bar{\mathbf{J}}_0) \big\|_2,
\end{equation}
where the joints $\mathbf{J}_j$ are regressed from each refined mesh with the wrist at $j = 0$ and bars denote the annotation.
MPVPE applies the same error to the $778$ MANO vertices, and Accel~\cite{kocabas2020vibe} to the second differences of the joints over consecutive frames.
Transfer measures the share of the canonical deformation $\mathbf{d}_k$ of Eqn.~\eqref{eq:delta_cano} that persists between adjacent frames and reaches $100\%$ when the deformation never changes.
With $\langle\cdot\rangle$ averaging over the clip, it is computed as
\begin{equation}
  \label{eq:app_tr}
  \mathrm{Tr.} = 1 - \big\langle \|\mathbf{d}_k - \mathbf{d}_{k+1}\| \big\rangle \,/\, \big\langle \|\mathbf{d}_{k+1}\| \big\rangle.
\end{equation}

\paragraph{Thermal contact.}
Following ContactOpt~\cite{grady2021contactopt,tse2022s2contact}, predicted contact consists of the object vertices within $2$\,mm of the hand and measured contact of those with a normalized thermal value above $0.4$.
We compare the two by precision, recall, and F1.
For the size-matched comparison, we enlarge the contact of the compared hand to as many object vertices as the lifted contact.

\paragraph{Contact estimators.}
To compare labels as supervision, we train contact estimators that differ only in their labels.
Each is a six-layer transformer~\cite{vaswani2017attention} that predicts per-vertex contact from geometric features of the rigid annotation and the object, such as signed distances and normals.
A label marks the hand vertices within $3$\,mm of the object on the rigid annotation or the lift, and a size-matched rigid label widens the rigid one per frame to the lifted contact count.
We train each estimator with three seeds on held-out lifted frames and score it on ContactPose by AP against the thermal contact on object vertices near the hand.

\section{Additional Quantitative Results}
\label{app:results}
This section provides additional quantitative results of our lift that extend the experiments in the main paper.
We first analyze the refined estimates and the learned head in \Cref{app:res_posthoc}.
We then test the lifted annotations on another dataset, under weaker friction, and as contact supervision in \Cref{app:res_plaus}.
Finally, we compare the lift with simpler alternatives in \Cref{app:res_alt}.

\subsection{Refining Estimated Hands}
\label{app:res_posthoc}
\paragraph{Consistency of the gains.}
Since \Cref{tab:posthoc} reports averages, we test whether its gains hold across clips by pairing each frame of the lift with the same frame of the estimate and of each pose refiner.
As shown in \Cref{tab:app_posthoc_ci}, the lift raises band conformity by $24$ to $28$ over every estimate and removes $25$ to $39$ deeply penetrating vertices per frame, with every interval excluding zero.
Meanwhile, the joint error grows by at most $0.31$ and stability stays close to that of the estimate.
Against the pose refiners, the lift lowers the joint error by $3.8$ to $20.2$ and makes the motion smoother.
Contact can thus be made consistent on every estimate without departing from its pose, leaving further stability to the learned head.

\begin{table}[H]
  \centering
  \caption{Differences between our lift and the estimate or each pose refiner of \Cref{tab:posthoc}, paired frame by frame on the test clips. Each entry is the lift minus the compared method with $\pm$ half the clip-clustered $95\%$ interval, so negative values favor the lift except for C2.}
  \label{tab:app_posthoc_ci}
  \vspace{-0.75em}
  \small
  \resizebox{\linewidth}{!}{%
  \begin{tabular}{l|l|N{-15.46}{0.07}|N{-1.32}{0.18}N{-38.8}{6.8}N{+27.9}{3.6}|N{-0.11}{0.10}N{+12.5}{1.7}|N{-13.54}{0.01}}
    \toprule
    & & \textbf{Hand} & \multicolumn{3}{c|}{\textbf{Contact}} & \multicolumn{2}{c|}{\textbf{Stability}} & \textbf{Temporal} \\
    \cmidrule(lr){3-3}\cmidrule(lr){4-6}\cmidrule(lr){7-8}\cmidrule(l){9-9}
    & & MPJPE & P.D. & $N_{\mathrm{in}>1}$ & C2 & S.D.@6 & F.E.@6 & Accel \\
    \textbf{Source} & \textbf{Compared with} & {\small (mm) $\downarrow$} & {\small (mm) $\downarrow$} & {\small (\#) $\downarrow$} & {\small (\%) $\uparrow$} & {\small (cm) $\downarrow$} & {\small (\%) $\downarrow$} & {\small (mm/f$^2$) $\downarrow$} \\
    \midrule
    \multirow{3}{*}{HOISDF~\cite{qi2024hoisdf}} & Estimate & \alignednum{$+0.17$}[0.07] & \alignednum{$-1.32$}[0.18] & \alignednum{$-38.8$}[6.8] & \alignednum{$+27.9$}[3.6] & \alignednum{$-0.11$}[0.10] & \alignednum{$-1.9$}[1.7] & \alignednum{$+0.03$}[0.01] \\
     & ContactOpt~\cite{grady2021contactopt} & \alignednum{$-5.00$}[0.87] & \alignednum{$-0.57$}[0.32] & \alignednum{$+0.8$}[6.2] & \alignednum{$+31.6$}[3.5] & \alignednum{$+0.15$}[0.17] & \alignednum{$+4.3$}[2.2] & \alignednum{$-13.54$}[1.89] \\
     & GeneOH Diffusion~\cite{liu2024geneoh} & \alignednum{$-6.62$}[1.07] & \alignednum{$-1.45$}[0.23] & \alignednum{$-47.0$}[7.9] & \alignednum{$+33.3$}[3.6] & \alignednum{$-0.04$}[0.11] & \alignednum{$+1.8$}[1.9] & \alignednum{$-16.95$}[2.21] \\
    \midrule
    \multirow{3}{*}{UniHOPE~\cite{wang2025unihope}} & Estimate & \alignednum{$+0.15$}[0.07] & \alignednum{$-0.67$}[0.14] & \alignednum{$-25.1$}[5.3] & \alignednum{$+27.5$}[3.5] & \alignednum{$-0.02$}[0.08] & \alignednum{$+0.3$}[1.1] & \alignednum{$+0.07$}[0.02] \\
     & ContactOpt~\cite{grady2021contactopt} & \alignednum{$-15.46$}[2.57] & \alignednum{$-0.40$}[0.34] & \alignednum{$+11.6$}[6.4] & \alignednum{$+34.2$}[3.5] & \alignednum{$+0.72$}[0.28] & \alignednum{$+12.5$}[3.5] & \alignednum{$-38.83$}[5.09] \\
     & GeneOH Diffusion~\cite{liu2024geneoh} & \alignednum{$-20.22$}[4.40] & \alignednum{$-0.13$}[0.29] & \alignednum{$-35.0$}[7.3] & \alignednum{$+36.1$}[3.6] & \alignednum{$-0.19$}[0.16] & \alignednum{$+1.1$}[1.9] & \alignednum{$-17.64$}[2.70] \\
    \midrule
    \multirow{3}{*}{VPHO~\cite{zhou2026vpho}} & Estimate & \alignednum{$+0.31$}[0.08] & \alignednum{$-1.39$}[0.18] & \alignednum{$-35.4$}[4.6] & \alignednum{$+24.4$}[2.9] & \alignednum{$-0.01$}[0.09] & \alignednum{$-0.1$}[1.3] & \alignednum{$+0.00$}[0.01] \\
     & ContactOpt~\cite{grady2021contactopt} & \alignednum{$-3.76$}[0.71] & \alignednum{$-0.63$}[0.31] & \alignednum{$-6.9$}[5.9] & \alignednum{$+27.7$}[3.2] & \alignednum{$+0.04$}[0.17] & \alignednum{$+0.7$}[2.7] & \alignednum{$-11.93$}[1.48] \\
     & GeneOH Diffusion~\cite{liu2024geneoh} & \alignednum{$-5.73$}[1.20] & \alignednum{$-1.81$}[0.25] & \alignednum{$-48.2$}[7.1] & \alignednum{$+31.9$}[2.9] & \alignednum{$+0.03$}[0.09] & \alignednum{$+2.3$}[1.8] & \alignednum{$-13.14$}[2.72] \\
    \bottomrule
  \end{tabular}}
\end{table}

\paragraph{Validation of the learned head.}
Since its guidance is tuned on $20$ test clips, we test the head on the other $60$ and against two controls: the same head trained on rigid annotations and as many attraction vertices chosen by proximity to the object.
As shown in \Cref{tab:app_learned_use}, guidance lowers F.E.@6 by $2.1$ to $4.7$ and raises C2 by $2.1$ to $2.9$ on every estimator, while the joint error grows by at most $0.16$.
Both controls leave a higher equilibrium residual on every estimator even at a fixed contact count, and the rigid-label head predicts the lifted contact less accurately with F1 lower by $0.07$ to $0.16$.
The guided lift also agrees best with the annotated contact on HOISDF and VPHO, reaching F1$_{\mathrm{GT}}$ of $0.35$ and $0.39$ against at most $0.31$ and $0.37$.
Lifted labels thus teach the head where to add stabilizing contact on unseen clips better than rigid labels or proximity.

\begin{table}[t]
  \centering
  \caption{Differences between the guided lift (Lift + Head) and the lift without guidance or with one of two controls, paired frame by frame. Rigid-Label Head is trained on rigid annotations, and Nearest Vertices adds as many attraction vertices as the head by proximity to the object. Each entry is \mbox{Lift + Head} minus the variant, so negative values favor it except for C2. Held-out Clips exclude the $20$ clips used for tuning.}
  \label{tab:app_learned_use}
  \vspace{-0.75em}
  \small
  \begin{tabularx}{\textwidth}{l|l|A{+0.16}{0.05}|A{+2.9}{1.1}|A{-0.31}{0.09}A{-4.66}{1.35}A{-4.10}{1.34}}
    \toprule
    & & \textbf{Hand} & \textbf{Contact} & \multicolumn{3}{c}{\textbf{Stability}} \\
    \cmidrule(lr){3-3}\cmidrule(lr){4-4}\cmidrule(l){5-7}
    & & MPJPE & C2 & S.D.@6 & F.E.@6 & F.E.$_{\mathrm{fix}}$ \\
    \textbf{Source} & \textbf{Compared with} & {\small (mm) $\downarrow$} & {\small (\%) $\uparrow$} & {\small (cm) $\downarrow$} & {\small (\%) $\downarrow$} & {\small (\%) $\downarrow$} \\
    \midrule
    \rowcolor{tabgray}\multicolumn{7}{c}{Held-out Clips} \\
    \midrule
    HOISDF~\cite{qi2024hoisdf} & Lift & \alignednum{$+0.16$}[0.05] & \alignednum{$+2.9$}[1.1] & \alignednum{$-0.31$}[0.09] & \alignednum{$-4.66$}[1.35] & \alignednum{$-4.10$}[1.34] \\
    UniHOPE~\cite{wang2025unihope} & Lift & \alignednum{$+0.06$}[0.03] & \alignednum{$+2.1$}[1.1] & \alignednum{$-0.08$}[0.06] & \alignednum{$-2.07$}[0.70] & \alignednum{$-2.77$}[1.06] \\
    VPHO~\cite{zhou2026vpho} & Lift & \alignednum{$+0.05$}[0.04] & \alignednum{$+2.3$}[0.8] & \alignednum{$-0.14$}[0.07] & \alignednum{$-2.25$}[1.19] & \alignednum{$-2.18$}[1.14] \\
    \midrule
    \rowcolor{tabgray}\multicolumn{7}{c}{All Clips} \\
    \midrule
    \multirow{3}{*}{HOISDF~\cite{qi2024hoisdf}} & Lift & \alignednum{$+0.17$}[0.04] & \alignednum{$+3.0$}[0.9] & \alignednum{$-0.32$}[0.08] & \alignednum{$-4.98$}[1.14] & \alignednum{$-4.43$}[1.14] \\
     & Lift + Rigid-Label Head & \alignednum{$+0.05$}[0.03] & \alignednum{$+0.2$}[0.4] & \alignednum{$-0.11$}[0.05] & \alignednum{$-1.60$}[0.61] & \alignednum{$-1.38$}[0.57] \\
     & Lift + Nearest Vertices & \alignednum{$+0.09$}[0.03] & \alignednum{$-2.3$}[0.5] & \alignednum{$-0.13$}[0.06] & \alignednum{$-0.98$}[0.52] & \alignednum{$-1.00$}[0.50] \\
    \cmidrule(lr){1-7}
    \multirow{3}{*}{UniHOPE~\cite{wang2025unihope}} & Lift & \alignednum{$+0.06$}[0.03] & \alignednum{$+2.1$}[1.0] & \alignednum{$-0.08$}[0.05] & \alignednum{$-2.11$}[0.57] & \alignednum{$-2.78$}[0.81] \\
     & Lift + Rigid-Label Head & \alignednum{$+0.02$}[0.02] & \alignednum{$+1.3$}[0.8] & \alignednum{$-0.00$}[0.05] & \alignednum{$-1.03$}[0.34] & \alignednum{$-1.69$}[0.73] \\
     & Lift + Nearest Vertices & \alignednum{$+0.02$}[0.02] & \alignednum{$-2.0$}[0.5] & \alignednum{$-0.05$}[0.05] & \alignednum{$-0.45$}[0.33] & \alignednum{$-0.74$}[0.50] \\
    \cmidrule(lr){1-7}
    \multirow{3}{*}{VPHO~\cite{zhou2026vpho}} & Lift & \alignednum{$+0.06$}[0.03] & \alignednum{$+2.5$}[0.6] & \alignednum{$-0.14$}[0.06] & \alignednum{$-2.23$}[0.90] & \alignednum{$-2.18$}[0.87] \\
     & Lift + Rigid-Label Head & \alignednum{$+0.04$}[0.02] & \alignednum{$+0.4$}[0.4] & \alignednum{$-0.05$}[0.04] & \alignednum{$-0.84$}[0.39] & \alignednum{$-0.81$}[0.40] \\
     & Lift + Nearest Vertices & \alignednum{$+0.04$}[0.02] & \alignednum{$-1.0$}[0.4] & \alignednum{$-0.04$}[0.05] & \alignednum{$-0.44$}[0.40] & \alignednum{$-0.44$}[0.40] \\
    \bottomrule
  \end{tabularx}
\end{table}

\paragraph{Skin deformation versus pose change.}
Since pose refiners move the hand far from the estimate, we ask whether ContactOpt~\cite{grady2021contactopt} matches the guided lift when limited to the same small pose change.
As shown in \Cref{tab:app_head_co}, the guided lift then exceeds it in band conformity by $27$ to $31$ and leaves $16$ to $28$ fewer deeply penetrating vertices per frame.
It is also more stable on HOISDF and VPHO with F.E.@6 lower by $4.5$ and $1.6$, and comparable on UniHOPE.
At its released setting ContactOpt moves the joints $3.7$ to $15.8$ further and is more stable only on UniHOPE, where the predicted object often lies out of reach.
Deforming the skin thus yields more consistent contact and grasps at least as stable as an equal change of pose.

\begin{table}[t]
  \centering
  \caption{Differences between the guided lift (Lift + Head) and ContactOpt~\cite{grady2021contactopt}, paired frame by frame. ContactOpt runs with a lower learning rate that matches the pose change of the lift (Matched Pose Change) or at its released setting. Each entry is \mbox{Lift + Head} minus ContactOpt averaged within each clip and then over clips, so it can differ from the frame means of \Cref{tab:posthoc}. Negative values favor the lift except for C2.}
  \label{tab:app_head_co}
  \vspace{-0.75em}
  \small
  \resizebox{\linewidth}{!}{%
  \begin{tabular}{l|l|N{-15.8}{2.6}|N{+39.8}{3.3}N{-28.3}{6.1}|N{-0.31}{0.30}N{+10.8}{3.6}}
    \toprule
    & & \textbf{Hand} & \multicolumn{2}{c|}{\textbf{Contact}} & \multicolumn{2}{c}{\textbf{Stability}} \\
    \cmidrule(lr){3-3}\cmidrule(lr){4-5}\cmidrule(l){6-7}
    & & MPJPE & C2 & $N_{\mathrm{in}>1}$ & S.D.@6 & F.E.@6 \\
    \textbf{Source} & \textbf{Compared with} & {\small (mm) $\downarrow$} & {\small (\%) $\uparrow$} & {\small (\#) $\downarrow$} & {\small (cm) $\downarrow$} & {\small (\%) $\downarrow$} \\
    \midrule
    \multirow{2}{*}{HOISDF~\cite{qi2024hoisdf}} & ContactOpt~\cite{grady2021contactopt} + Matched Pose Change & \alignednum{$-0.1$}[0.1] & \alignednum{$+29.5$}[2.9] & \alignednum{$-28.3$}[4.7] & \alignednum{$-0.31$}[0.12] & \alignednum{$-4.5$}[2.0] \\
     & ContactOpt~\cite{grady2021contactopt} + Released Setting & \alignednum{$-4.9$}[0.9] & \alignednum{$+35.2$}[3.0] & \alignednum{$+0.0$}[6.1] & \alignednum{$-0.15$}[0.17] & \alignednum{$-0.7$}[2.3] \\
    \midrule
    \multirow{2}{*}{UniHOPE~\cite{wang2025unihope}} & ContactOpt~\cite{grady2021contactopt} + Matched Pose Change & \alignednum{$-0.2$}[0.2] & \alignednum{$+30.9$}[3.3] & \alignednum{$-16.0$}[3.3] & \alignednum{$+0.01$}[0.08] & \alignednum{$-1.3$}[1.4] \\
     & ContactOpt~\cite{grady2021contactopt} + Released Setting & \alignednum{$-15.8$}[2.6] & \alignednum{$+39.8$}[3.2] & \alignednum{$+10.4$}[5.7] & \alignednum{$+0.68$}[0.30] & \alignednum{$+10.8$}[3.6] \\
    \midrule
    \multirow{2}{*}{VPHO~\cite{zhou2026vpho}} & ContactOpt~\cite{grady2021contactopt} + Matched Pose Change & \alignednum{$-0.1$}[0.1] & \alignednum{$+26.6$}[2.4] & \alignednum{$-24.4$}[2.6] & \alignednum{$-0.06$}[0.09] & \alignednum{$-1.6$}[1.3] \\
     & ContactOpt~\cite{grady2021contactopt} + Released Setting & \alignednum{$-3.7$}[0.7] & \alignednum{$+30.4$}[2.7] & \alignednum{$-6.5$}[5.7] & \alignednum{$-0.09$}[0.16] & \alignednum{$-1.6$}[2.3] \\
    \bottomrule
  \end{tabular}}
\end{table}

\subsection{Lifting GT Annotations}
\label{app:res_plaus}
\begin{wraptable}[8]{r}{0.55\linewidth}%
  \vspace{-\intextsep}
  \centering
  \caption{The HO3D~\cite{hampali2020honnotate} annotations lifted stage by stage, marked as in \Cref{tab:dataset_plausibility}. Seating removes most deep penetration, while skin deformation adds most of the contact and stability.}
  \label{tab:app_ho3d_stages}
  \vspace{-0.75em}
  \small
  \resizebox{\linewidth}{!}{%
  \begin{tabular}{l|cc|c|cc|cccc}
    \toprule
    & \multicolumn{2}{c|}{} & \textbf{Contact} & \multicolumn{2}{c|}{\textbf{Penetration}} & \multicolumn{4}{c}{\textbf{Stability}} \\
    \cmidrule(lr){4-4}\cmidrule(lr){5-6}\cmidrule(l){7-10}
    & &
      & C2
      & P.D. & $N_{\mathrm{in}>1}$
      & S.D.@1 & S.D.@6 & F.E.@6 & F.E.$_{\mathrm{fix}}$ \\
    \textbf{Dataset} & \textbf{Seating} & \textbf{Skin}
      & {\small (\%) $\uparrow$}
      & {\small (mm) $\downarrow$} & {\small (\#) $\downarrow$}
      & {\small (cm) $\downarrow$} & {\small (cm) $\downarrow$} & {\small (\%) $\downarrow$} & {\small (\%) $\downarrow$} \\
    \midrule
      \multirow{3}{*}{HO3D~\cite{hampali2020honnotate}} &  &  & \underline{18.2} & 5.19 & 20.2 & 1.52 & 1.29 & 7.1 & 6.7 \\
       & \cmark &  & 17.7 & \textbf{4.05} & \underline{5.7} & \underline{1.44} & \underline{1.24} & \underline{6.1} & \underline{6.1} \\
       & \gc\cmark & \gc\cmark & \gc \textbf{89.2} & \gc \underline{4.42} & \gc \textbf{1.9} & \gc \textbf{1.25} & \gc \textbf{1.08} & \gc \textbf{2.2} & \gc \textbf{2.9} \\
    \bottomrule
  \end{tabular}}
\end{wraptable}
\paragraph{Generalization to HO3D.}
To test the lift on another dataset, we apply its stages to the HO3Dv2 annotations~\cite{hampali2020honnotate}.
As shown in \Cref{tab:app_ho3d_stages}, the full lift raises C2 from $18$ to $89$, reduces $N_{\mathrm{in}>1}$ from $20.2$ to $1.9$, and lowers F.E.@6 from $7.1$ to $2.2$.
Seating alone changes neither C2 nor F.E.@6 significantly, so skin deformation drives both gains as on DexYCB and OakInk.

\begin{table}[t]
  \centering
  \caption{Closed-form and pose-space alternatives to the lift on the estimated inputs of \Cref{tab:posthoc}, all starting from the seated hand. The contact-band rule is also smoothed $100$ times more ($\dagger$) or filtered over time by a Gaussian of $\sigma = 2$ frames ($\ddagger$). Skin columns are measured against the seated hand. Only the lift reaches high conformity while flipping almost no faces and changing the least between frames.}
  \label{tab:app_posthoc_cf}
  \vspace{-0.75em}
  \small
  \resizebox{\linewidth}{!}{%
  \begin{tabular}{l|l|cccc|cc|ccc}
    \toprule
    & & \multicolumn{4}{c|}{\textbf{Contact}} & \multicolumn{2}{c|}{\textbf{Stability}} & \multicolumn{3}{c}{\textbf{Skin}} \\
    \cmidrule(lr){3-6}\cmidrule(lr){7-8}\cmidrule(l){9-11}
    & & P.D. & I.V. & $N_{\mathrm{in}>1}$ & C2 & S.D.@6 & F.E.@6 & Stretch & Flips & Change \\
    \textbf{Source} & \textbf{Method} & {\small (mm) $\downarrow$} & {\small (cm$^3$) $\downarrow$} & {\small (\#) $\downarrow$} & {\small (\%) $\uparrow$} & {\small (cm) $\downarrow$} & {\small (\%) $\downarrow$} & {\small (\%) $\downarrow$} & {\small (\#) $\downarrow$} & {\small (mm) $\downarrow$} \\
    \midrule
    \multirow{7}{*}{HOISDF~\cite{qi2024hoisdf}} & Estimate & 6.44 & 7.57 & 57.0 & 20.7 & 2.31 & 18.0 & -- & -- & -- \\
     & Clip to $\epsilon$ & 4.54 & 4.77 & 0.0 & 20.2 & 2.48 & 23.4 & 26 & 1.06 & 1.15 \\
     & Contact-Band Rule & 4.70 & 6.04 & 0.0 & 51.1 & 2.16 & 16.4 & 41 & 1.83 & 2.44 \\
     & Contact-Band Rule$^{\dagger}$ & 4.80 & 5.49 & 0.0 & 29.2 & 2.25 & 18.6 & 24 & 0.79 & 1.96 \\
     & Contact-Band Rule$^{\ddagger}$ & 4.56 & 5.33 & 0.0 & 32.7 & 2.25 & 18.7 & 40 & 1.57 & 1.07 \\
     & Pose Space & 5.27 & 5.33 & 21.6 & 20.9 & 2.32 & 18.3 & 9 & 0.01 & 0.99 \\
     & \gc Lift (ours) & \gc 5.13 & \gc 5.98 & \gc 18.2 & \gc 48.6 & \gc 2.19 & \gc 16.1 & \gc 27 & \gc 0.10 & \gc 0.90 \\
    \midrule
    \multirow{7}{*}{UniHOPE~\cite{wang2025unihope}} & Estimate & 5.06 & 6.28 & 48.4 & 20.0 & 5.29 & 46.2 & -- & -- & -- \\
     & Clip to $\epsilon$ & 3.77 & 4.81 & 0.0 & 23.7 & 5.42 & 50.6 & 32 & 1.51 & 2.43 \\
     & Contact-Band Rule & 3.88 & 5.64 & 0.0 & 48.4 & 5.25 & 46.8 & 38 & 2.60 & 3.69 \\
     & Contact-Band Rule$^{\dagger}$ & 3.87 & 5.23 & 0.0 & 26.4 & 5.35 & 48.3 & 26 & 1.10 & 3.43 \\
     & Contact-Band Rule$^{\ddagger}$ & 3.75 & 4.99 & 0.0 & 27.0 & 5.35 & 49.5 & 41 & 2.03 & 2.00 \\
     & Pose Space & 4.42 & 5.05 & 26.0 & 20.1 & 5.40 & 48.3 & 8 & 0.04 & 1.58 \\
     & \gc Lift (ours) & \gc 4.39 & \gc 5.47 & \gc 23.4 & \gc 47.5 & \gc 5.27 & \gc 46.5 & \gc 19 & \gc 0.09 & \gc 1.37 \\
    \midrule
    \multirow{7}{*}{VPHO~\cite{zhou2026vpho}} & Estimate & 6.55 & 8.23 & 48.4 & 22.3 & 1.54 & 9.3 & -- & -- & -- \\
     & Clip to $\epsilon$ & 4.55 & 4.80 & 0.0 & 19.4 & 1.81 & 15.7 & 22 & 0.64 & 0.92 \\
     & Contact-Band Rule & 4.79 & 6.65 & 0.0 & 55.0 & 1.49 & 9.3 & 41 & 1.14 & 2.45 \\
     & Contact-Band Rule$^{\dagger}$ & 4.90 & 5.92 & 0.0 & 31.6 & 1.62 & 11.2 & 22 & 0.49 & 1.91 \\
     & Contact-Band Rule$^{\ddagger}$ & 4.69 & 5.74 & 0.0 & 33.7 & 1.61 & 11.3 & 39 & 1.06 & 0.94 \\
     & Pose Space & 5.10 & 5.24 & 15.9 & 19.4 & 1.69 & 12.4 & 9 & 0.01 & 0.87 \\
     & \gc Lift (ours) & \gc 5.17 & \gc 6.24 & \gc 13.0 & \gc 46.7 & \gc 1.53 & \gc 9.3 & \gc 29 & \gc 0.08 & \gc 0.85 \\
    \bottomrule
  \end{tabular}}
\end{table}

\paragraph{Weaker friction and contact supervision.}
Finally, we ask whether the lifted grasps hold under weaker friction and whether the lifted annotations make better supervision.
With the friction coefficient of the simulation in \Cref{tab:dataset_plausibility} lowered from $3$ to $0.5$ and $1$, lifting still reduces S.D.@6 by $0.73$ and $0.28$ on DexYCB and by $0.27$ and $0.11$ on OakInk with every interval excluding zero.
Transformer contact estimators~\cite{vaswani2017attention} trained on lifted labels reach an AP of $0.529$ against thermal contact, compared with $0.497$ for rigid labels and only $0.499$ for size-matched rigid labels.
This gain of $0.032$ holds for every seed with a $95\%$ interval from $0.030$ to $0.035$.
It stays between $0.030$ and $0.036$ when training on $5{,}000$ to all $50{,}600$ instants and reaches $0.034$ without the signed-distance and band inputs.
The lifted grasps thus do not rely on high friction, and the lifted labels teach where contact lies rather than merely how much.

\subsection{Simpler Alternatives}
\label{app:res_alt}
\paragraph{Closed-form and pose-space alternatives.}
We ask whether a simpler correction of the seated hand could replace the lift on the estimated inputs of \Cref{tab:posthoc}.
The closed forms keep the seated pose and clip penetration to $\epsilon$, which removes every vertex deeper than $1$\,mm even where the lift exempts deep source penetration.
The contact-band rule further applies a Laplacian edit~\cite{sorkine2004laplacian} that projects the contact band onto the object and smooths the rest of the surface.
The pose-space variant instead keeps each skinning segment rigid and optimizes only the joints, the wrist, and one translation per segment under the same temporal prior.
We also measure Stretch as the $99$th percentile of the relative change in edge length from the seated skin and Flips as the number of faces whose normal turns by more than $90^{\circ}$.
As shown in \Cref{tab:app_posthoc_cf}, clipping alone leaves conformity at the source level and makes every estimate less stable.
The rule matches the conformity of the lift but stretches the skin up to twice as much and flips over ten times as many faces.
Smoothing it $100$ times more brings its stretch close to that of the lift but lowers conformity to about $30$ and leaves F.E.@6 $1.8$ to $2.5$ above the lift.
The pose-space variant distorts the skin least but cannot bend it onto the object, so its conformity stays near the source level and its grasps are less stable.
Among these alternatives, only the lift reaches high conformity while flipping almost no faces.

\newpage
\vspace*{-\topskip}%
\begin{wrapfigure}{r}{0.6\linewidth}
  \vspace{-\intextsep}
  \centering
  \begin{minipage}[b]{0.48\linewidth}\includegraphics[width=\linewidth]{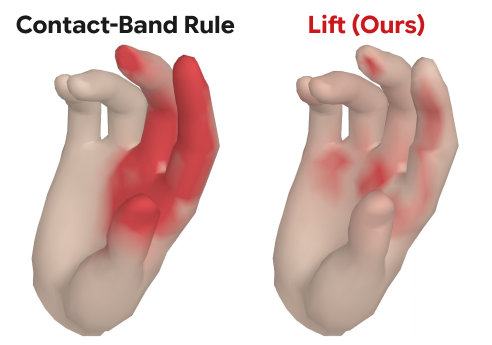}\end{minipage}\hfill
  \begin{minipage}[b]{0.48\linewidth}\includegraphics[width=\linewidth]{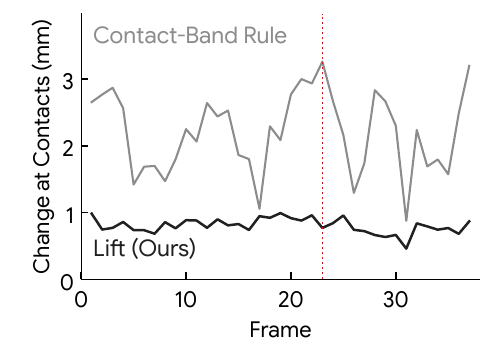}\end{minipage}
  \caption{Change between consecutive frames for the contact-band rule and the lift on a typical VPHO~\cite{zhou2026vpho} clip. The hands show it at the frame marked by the dashed line, from skin color for none to \textcolor[HTML]{D11417}{red} for $2$\,mm or more. The plot traces its mean over the contact vertices. The rule fluctuates while the lift stays low and steady.}
  \label{fig:app_cf_flicker}
  \vspace{-\baselineskip}
\end{wrapfigure}
\paragraph{Change between frames.}
Since the rule treats each frame independently, we also measure Change, the mean change of displacement on the contact vertices between consecutive frames.
It is two to three times larger for the rule than for the lift, whose frames share the temporal prior.
\Cref{fig:app_cf_flicker} shows the rule fluctuating on a VPHO clip while the lift stays low and steady.
Filtering the rule over time steadies it but lowers conformity to about $30$.
A closed form thus gains conformity only with distorted and flickering skin, and a rigid correction cannot gain it at all.

\begin{wraptable}{r}{0.4\linewidth}
  \vspace{-\intextsep}
  \centering
  \caption{Agreement with observed depth on the contact rim of DexYCB test frames. $|r|$ is the mean absolute residual and $\Delta$ its difference from the lift with $\pm$ half the $95\%$ interval, so positive values favor the lift.}
  \label{tab:app_depth}
  \vspace{-0.75em}
  \small
  \begin{tabularx}{\linewidth}{l|Y|A{-0.16}{0.34}}
    \toprule
    \textbf{Method} & $|r|$ {\small (mm) $\downarrow$} & $\Delta$ {\small (mm)} \\
    \midrule
    Rigid Annotation (GT) & 4.75 & \alignednum{$-0.16$}[0.34] \\
    Seating Alone & 5.26 & \alignednum{$+0.36$}[0.17] \\
    Pose Space & 5.21 & \alignednum{$+0.30$}[0.16] \\
    Contact-Band Rule & 6.10 & \alignednum{$+1.20$}[0.62] \\
    \rowcolor{tabgray} Lift (ours) & 4.91 & --- \\
    \bottomrule
  \end{tabularx}
\end{wraptable}
\paragraph{Agreement with observed depth.}
Since no loss of the lift uses the DexYCB depth images, we finally test whether conforming the skin moves it away from where they observed it.
We render each hand with its object into the color camera, remove the camera offset of each frame, and compare its depth with the observation on the contact rim, the pixels where every compared hand is the nearest surface.
As shown in \Cref{tab:app_depth}, the lift stays within $0.16$ of the rigid annotation even though that annotation was fitted to these images.
Seating alone and the pose-space variant fit $0.3$ to $0.4$ worse, and the contact-band rule fits $1.2$ worse.
The lift thus makes contact consistent while keeping the skin on the observed surface, which the simpler alternatives do not.

\section{Limitations and Future Work}
\label{app:limitations}

\paragraph{Limitations.}
Although the lift makes contact consistent on annotated and estimated hands, a few limitations remain.
First, the skin displacement is optimized for this consistency rather than tissue mechanics and need not match a measured indentation.
Second, the lift keeps the predicted object fixed and the hand near its estimate to preserve pose accuracy.
A grasp may thus stay open when the predicted object lies out of reach, as often happens for some estimators (\eg, UniHOPE~\cite{wang2025unihope}).
Third, the learned contact head reads the features of one estimator and must be retrained for each new one, albeit without manual labeling.
Addressing these limitations would make the lifted hands more physically faithful and readily applicable to any estimator.

\paragraph{Future work.}
Our work opens several directions for future research.
First, calibrating the skin displacement on datasets with pressure measurements~\cite{grady2022pressurevision,zhao2025egopressure} could let its magnitude reflect physical indentation.
Second, refining the object pose along with the hand could close the grasps that stay open when the predicted object lies out of reach.
Third, training estimators on the lifted annotations could yield contact-consistent hands directly from images without a separate head or test-time optimization.
Progress in these directions would take hand-object reconstruction further beyond rigid hands.